\documentclass[final]{cvpr}

\usepackage{times}
\usepackage{epsfig}
\usepackage{graphicx}
\usepackage{amsmath}
\usepackage{amssymb}
\usepackage{verbatim}
\usepackage{color}
\usepackage{xcolor}

\usepackage{subcaption}
\usepackage{wrapfig}
\usepackage{color,colortbl}
\usepackage{tabu}
\usepackage{arydshln}
\usepackage{diagbox}
\usepackage{pifont}
\usepackage{makecell}
\usepackage{multirow}
\usepackage{overpic}
\usepackage{amsmath}
\usepackage{bm}
\usepackage{algorithm}
\usepackage{algorithmic}
\usepackage{pifont}
\usepackage{rotating}

\usepackage[toc,page]{appendix}

\definecolor{myred}{RGB}{255,0,0}
\definecolor{mygreen}{RGB}{0,255,0}
\definecolor{myblue}{RGB}{0,0,255}

\newcommand{\redbf}[1]{\textbf{\textcolor{red}{#1}}}
\newcommand{\bluebf}[1]{\textbf{\textcolor{blue}{#1}}}

\usepackage{xcolor}
\usepackage{overpic}

\newcommand{\citecomment}[1]{[\textcolor{green}{$\ast$}]{}}

\graphicspath{{appendix/}}

\newcommand{\ecite}[1]{\etal\cite{#1}}

\usepackage[pagebackref=true,breaklinks=true,colorlinks,bookmarks=false]{hyperref}

\usepackage{cleveref}
\crefformat{section}{\S#2#1#3}
\crefformat{subsection}{\S#2#1#3}
\crefformat{subsubsection}{\S#2#1#3}
\crefformat{figure}{Figure~#2#1#3}
\crefformat{table}{Table~#2#1#3}

\def\cvprPaperID{3475}
\def\confYear{CVPR 2021}

\begin{document}

\title{Temporal Modulation Network for Controllable Space-Time Video Super-Resolution}

\author{
	Gang Xu$^{1}$
	\quad Jun Xu$^{2}$\thanks{Jun Xu is the corresponding author (email:\ nankaimathxujun@gmail.com).\ This work is supported by National Natural Science Foundation of China under Grant 62002176 and 61922046.
	}
	\quad Zhen Li$^{1}$
	\quad Liang Wang$^{3}$
	\quad Xing Sun$^{4}$
	\quad Ming-Ming Cheng$^{1}$\\
	$^{1}$ College of Computer Science, Nankai University, Tianjin, China\\
	$^{2}$ School of Statistics and Data Science, Nankai University, Tianjin, China\\
	$^{3}$  National Lab of Pattern Recognition, Institute of Automation, CAS, Beijing, China\\
	$^{4}$ Youtu Lab., Tencent, Shanghai, China\\
}

\maketitle

\pagestyle{empty}  
\thispagestyle{empty} 

\begin{abstract}
Space-time video super-resolution (STVSR) aims to increase the spatial and temporal resolutions of low-resolution and low-frame-rate videos.
Recently, deformable convolution based methods have achieved promising STVSR performance, but they could only infer the intermediate frame pre-defined in the training stage.\
Besides, these methods undervalued the short-term motion cues among adjacent frames.\
In this paper, we propose a Temporal Modulation Network (TMNet) to interpolate arbitrary intermediate frame(s) with accurate high-resolution reconstruction.\
Specifically, we propose a Temporal Modulation Block (TMB) to modulate deformable convolution kernels for controllable feature interpolation.\
To well exploit the temporal information, we propose a Locally-temporal Feature Comparison (LFC) module, along with the Bi-directional Deformable ConvLSTM, to extract short-term and long-term motion cues in videos.\
Experiments on three benchmark datasets demonstrate that our TMNet outperforms previous STVSR methods.\
The code is available at \url{https://github.com/CS-GangXu/TMNet}.
\end{abstract}

\section{Introduction}

Nowadays, flat-panel displays using liquid-crystal display (LCD) or light-emitting diode (LED) technologies can broadcast Ultra High Definition Television (UHD TV) videos with 4K ($3840\times2160$) or 8K ($7680\times4320$) full-color pixels, at the frame rate of 120 frames per second (FPS) or 240 FPS~\cite{etsi2019}.
However, currently available videos are commonly in Full High Definition (FHD) with a resolution of 2K ($1920\times1080$) at 30 FPS~\cite{xiang2020zooming}.
To broadcast FHD videos on UHD TVs, it is necessary to increase their space-time resolutions comfortably with the broadcasting standard of UHD TVs.
Although it is possible to increase the spatial resolution of videos frame-by-frame via single image super-resolution methods~\cite{dong2015image,lim2017enhanced}, the perceptual quality of the enhanced videos would be deteriorated by temporal distortion~\cite{kim2020fisr}.
To this end, the space-time video super-resolution (STVSR) methods~\cite{shechtman2002increasing,xiang2020zooming} are developed to simultaneously increase the spatial and temporal resolutions of low-frame-rate and low-resolution videos.
\begin{figure}[t]
\vspace{-0mm}
\includegraphics[width=8.3cm]{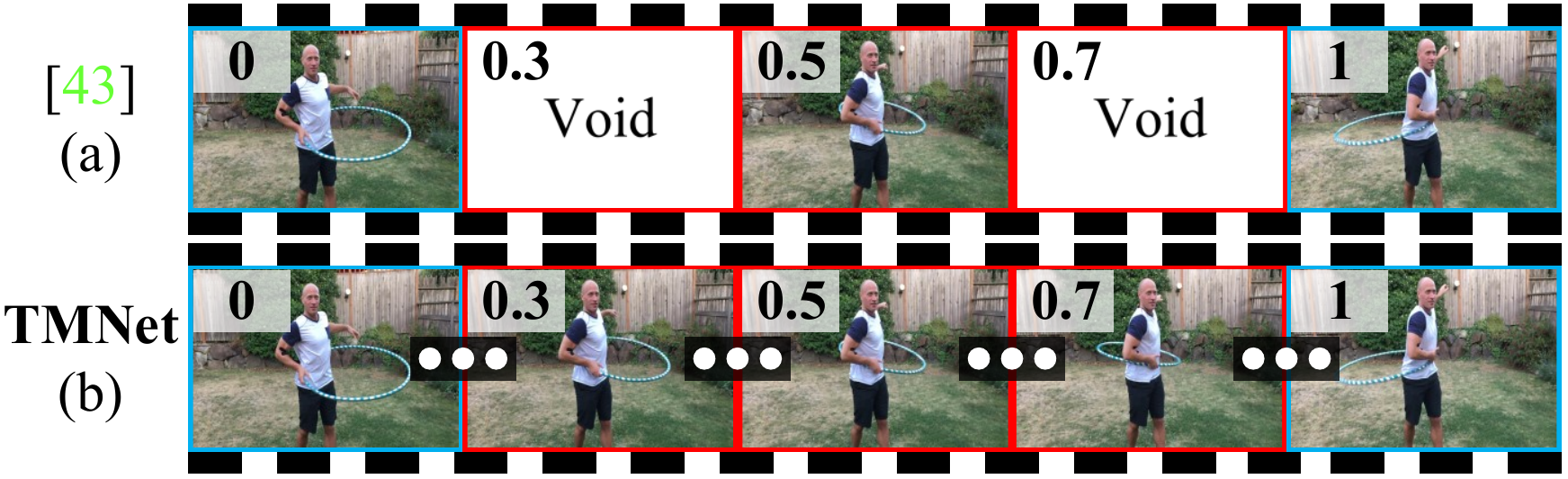}
\vspace{-7mm}
\caption{\textbf{Flexible STVSR performance by our TMNet}.
Given input frames at moments $\bm{0}$ (begin) and $\bm{1}$ (end),~\cite{xiang2020zooming} could only interpolate pre-defined intermediate frame at moment $\bm{0.5}$ (a), while our TMNet can generate intermediate frames at arbitrary moments (e.g., $\bm{0.3}$, $\bm{0.5}$, $\bm{0.7}$) (b).}
\label{fig:example}
\vspace{-6mm}
\end{figure}

Previous model-based STVSR methods~\cite{shechtman2002increasing,shechtman2005space,shahar2011space} rely heavily on precise spatial and temporal registration~\cite{tsai1984MultiframeIR}, and would produce inferior reconstruction results when the registration is inaccurate.\ Besides, they usually require huge computational costs on solving complex optimization problems, resulting in low inference efficiency~\cite{mudenagudi2010space,li2015space}.
Later, deep convolutional neural networks~\cite{resnet,lstm,convlstm, wu2021jcs} have been widely employed in video restoration tasks such as video super-resolution (VSR)~\cite{caballero2017real, tian2020tdan}, video frame interpretation (VFI)~\cite{niklaus2020softmax, xue2019video, bao2019depth},
and the more challenging STVSR~\cite{kim2020fisr,xiang2020zooming}.
A straightforward solution for STVSR is to perform VFI and VSR successively on low-resolution and low-frame-rate videos, to increase their spatial resolutions and frame rates~\cite{xiang2020zooming}.
However, these two-stage methods ignore the inherent correlation between temporal and spatial dimensions.
That is, the videos with high-resolution frames contain richer details on moving object(s) and background, while those in high-frame-rate provide finer pixel alignment between adjacent frames~\cite{haris2020space}.
Therefore, these two-stage STVSR methods would suffer from the temporal inconsistency problem~\cite{xiang2020zooming} and produce artifacts, e.g., ``the attentional blink phenomenon''~\cite{tang2020neural} on STVSR.

To well exploit the correlation between the temporal and spatial dimensions in videos, several one-stage STVSR methods~\cite{haris2020space,kim2020fisr,xiang2020zooming} have been proposed to simultaneously perform VFI and VSR reconstruction on low-frame-rate and low-resolution videos.
The work of STARnet~\cite{haris2020space} estimates the motion cues with an additional optical flow branch~\cite{flownet}, and performs feature warping of two adjacent frames to interpolate the intermediate frame.
But this flow-based method~\cite{haris2020space} needs to learn an extra branch for optical flow estimation, which consumes expensive costs on computation and memory.
To alleviate this problem, Xiang~\ecite{xiang2020zooming} employed the deformable convolution backbone~\cite{wang2019edvr}, and directly performed STVSR on the feature space.\
Though with promising performance, current STVSR networks could only generate the intermediate frames pre-defined in the network architecture, and thus are restricted to highly-controlled application scenarios with fixed frame-rate videos.
However, in many commercial scenarios, such as sports events, it is very common for the user to flexibly adjust the intermediate video frames for better visualization.
Thus, it is necessary to develop controllable STVSR methods for smooth motion synthesizing.

To fulfill the versatile requirements of broadcasting scenarios, in this paper, we propose a Temporal Modulation Network (TMNet) to interpolate an arbitrary number of intermediate frames for STVSR, as shown in Figure~\ref{fig:example}.
But current deformable convolution based methods~\cite{xiang2020zooming} could only generate pre-defined intermediate frame(s).
To tackle this problem, we propose a Temporal Modulation Block (TMB) to incorporate motion cues into the feature interpolation of intermediate frames.
Specifically, we first estimate the motion between two adjacent frames under the deformable convolution framework~\cite{wang2019edvr}, and learn controllable interpolation at an arbitrary moment defined by a temporal parameter.
In addition, we also propose a Locally-temporal Feature Comparison module to fuse multi-frame features for effective spatial alignment and feature warping, and a globally-temporal feature fusion to explore the long-term variations of the whole video. 
This two-stage temporal feature fusion scheme accurately interpolates the intermediate frames for STVSR.
Extensive experiments on three benchmarks~\cite{xue2019video, vid4, su2017deep} demonstrate that our TMNet is able to interpolate an arbitrary number of intermediate frames, and achieves state-of-the-art performance on STVSR.

The contribution of this work are three-fold:
\begin{itemize}
\vspace{-1mm}
   \item \textbf{We propose a Temporal Modulation Network (TMNet) to perform controllable interpolation of arbitrary frame-rates} for flexible STVSR performance.
   This is achieved by our Temporal Modulation Block under the deformable convolution framework.
\vspace{-1mm} 
   \item \textbf{We present a two-stage temporal feature fusion scheme for effective STVSR}.
   Specifically, we propose a locally-temporal feature comparison module to exploit the short-term motion cues of adjacent frames, and perform globally-temporal feature fusion by exploring the long-term variations over the whole video.
   
   \vspace{-1mm} 
   \item Experiments on three benchmarks show that \textbf{our TMNet is able to perform controllable frame interpolation at arbitrary frame-rate}, and outperforms state-of-the-art STVSR methods.
\end{itemize}

\section{Related Work}
\label{sec:related}

\noindent
\textbf{Video frame interpolation} (VFI) aims to synthesize new intermediate frames between adjacent frames~\cite{jiang2018super,bao2019memc,lee2020adacof}.
Early VFI methods mainly resort to optical flow techniques for motion estimation~\cite{jiang2018super,bao2019memc,niklaus2020softmax}.
Jiang~\ecite{jiang2018super} modeled motion interpretation for arbitrary frame-rate VFI.
Niklaus~\ecite{niklaus2018context} warped the input frames with contextual information, and interpolated context-aware intermediate frames.
Bao \etal employed motion estimation and compensation for VFI in~\cite{bao2019memc}, and obtained improved performance by further exploring the depth information~\cite{bao2019depth}.
Niklaus~\ecite{niklaus2020softmax} tackled the conflicts of mapping multiple pixels to the same location in VFI by softmax splatting.
However, these optical flow based methods need huge computational costs on motion estimation.\
Therefore, recently researchers exploited to learn spatially-adaptive convolution kernels~\cite{niklaus2017video} or deformable ones for VFI~\cite{lee2020adacof}.
\begin{figure*}[t]
\vspace{-1mm}
\centering
\includegraphics[width=17cm]{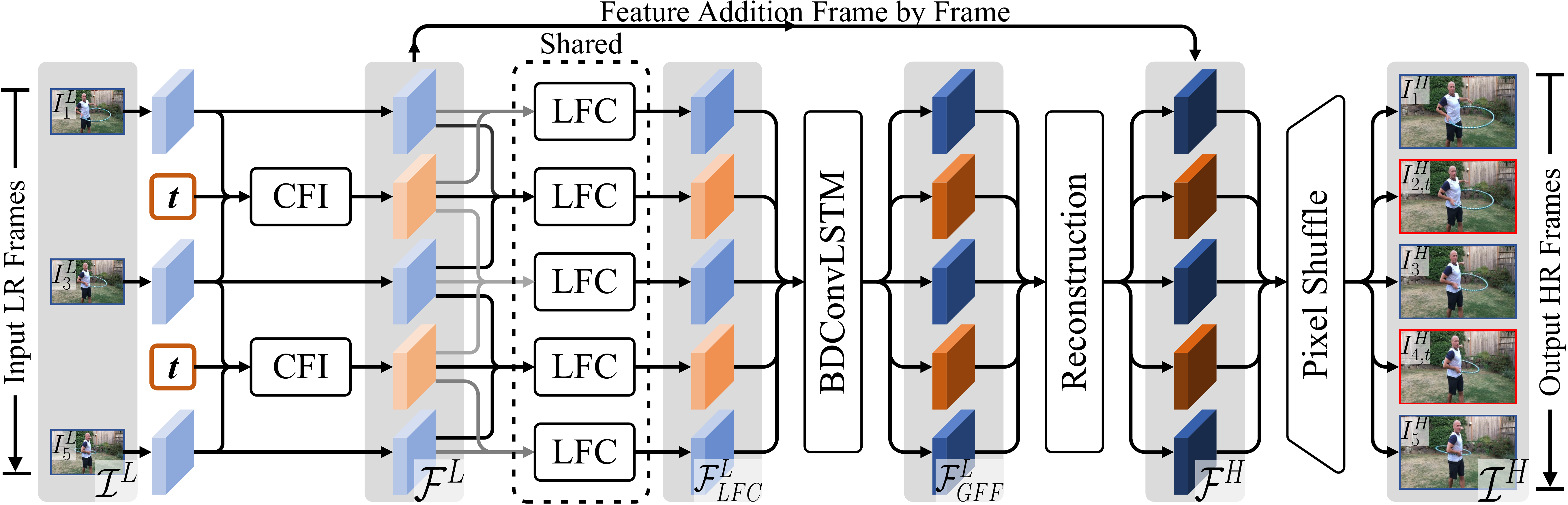} 
\vspace{-4mm}
\caption{\textbf{Overview of our Temporal Modulation Network (TMNet) for STVSR}.
Given the input low-frame-rate and low-resolution video $\mathcal{I}^{L}$, we first extract initial features and perform Controllable Feature Interpolation (CFI, implemented by our Temporal Modulation Block) for the intermediate frame at an arbitrary moment $t\in(0,1)$.
Then, we feed the obtained feature maps $\mathcal{F}^{L}$ into a two-stage temporal feature fusion scheme.
For short-term motion consistency, the feature maps $\mathcal{F}^{L}$ are refined to $\mathcal{F}_{LFC}^{L}$ by our Locally-temporal Feature Comparison (LFC) module.
To exploit long-term motion cues, the feature maps $\mathcal{F}_{LFC}^{L}$ are further improved to $\mathcal{F}_{GFF}^{L}$ by globally-temporal feature fusion (GFF), implemented by Bi-directional Deformable ConvLSTM (BDConvLSTM)~\cite{xiang2020zooming}.
Finally, we employ 40 residual blocks to reconstruct high-resolution feature maps $\mathcal{F}^{H}$, and two Pixel-Shuffle layers to output the high-frame-rate and high-resolution video $\mathcal{I}^{H}$.
}
\label{method}
\vspace{-4mm}
\end{figure*}

\noindent
\textbf{Video super-resolution} (VSR) is the task of increasing the spatial resolutions of low-resolution (LR) videos~\cite{jo2018deep,wang2019edvr,tian2020tdan}.
Existing VSR methods~\cite{jo2018deep,wang2019edvr,tian2020tdan} mainly aggregate spatial information of multiple frames for high-resolution (HR) reconstruction, with the help of optical flow techniques~\cite{flownet}.
Jo~\ecite{jo2018deep} generated dynamic upsampling filters to enhance the LR frames with residual learning~\cite{resnet}.
Wang~\ecite{wang2019edvr} proposed the Pyramid, Cascading and Deformable (PCD) module to perform frame alignment, and then fused multiple frames into a single one by spatial and temporal attention.
Haris~\ecite{haris2019recurrent} designed an iterative refinement framework by integrating the spatial and temporal contexts of multiple frames.
Tian~\ecite{tian2020tdan} utilized the learned sampling offsets of deformable convolution kernels to align the supporting frames with the reference ones, which are both used to reconstruct the HR frames.

\noindent
\textbf{Space-time video super-resolution} (STVSR) aims to increase the spatial and temporal dimensions of the low-frame-rate and low-resolution videos~\cite{kim2020fisr,haris2020space,xiang2020zooming}.
Shechtman~\ecite{shechtman2002increasing} tackled the STVSR problem by employing a directional space-time smoothness regularization on the HR video reconstruction problem.
Mudenagudi~\ecite{mudenagudi2010space} formulated their STVSR method under the Markov Random Field framework~\cite{geman1984stochastic}.
STARnet~\cite{haris2020space} leveraged inherent motion relationship between spatial and temporal dimensions with an extra optical flow branch~\cite{flownet}, and perform feature warping of two adjacent frames to interpolate the intermediate frame.
%
Xiang~\ecite{xiang2020zooming} developed a unified framework to interpolate the multi-frame features via PCD alignment modules~\cite{wang2019edvr}, the intermediate features by bidirectional deformable ConvLSTM~\cite{convlstm}, and finally performed STVSR by multi-frame feature fusion.
In this work, our goal is to develop a temporally controllable network for powerful and flexible STVSR.
Though built upon~\cite{xiang2020zooming}, our TMNet arrives at better performance on benchmark datasets, owing to the proposed locally temporal feature comparison module.

\noindent
\textbf{Modulation networks}.
%
Recently, researchers proposed to control the restoration intensity of the main network by additional modulation branches~\cite{he2019modulating,wang2019cfsnet,wang2019deep,he2019multi}.
These modulation networks are trained to trade-off the restoration quality and flexibility, which are controlled by hyper-parameters.
%
He~\ecite{he2019modulating} put feature modulation filters after each convolution layer to modulate the output according to user's preference.
Later, He~\ecite{he2019multi} expanded this design to multiple dimensions, and modulated the output according to the levels of multiple degradation types.
Wang~\ecite{wang2019cfsnet} learned the features from tuning blocks and residual ones with different objectives, to control the trade-off between noise reduction and detail preservation.
In this work, we consider the modulation on temporal dimension, instead of on restoration intensity as in these modulation networks.
As far as we know, our work is among the first to implement temporal modulation in the STVSR problem.
As will be shown in~\S\ref{sec:experiment}, our TMNet can explore the potential of temporal modulation for controllable STVSR.
\begin{figure*}[t]
    \centering
	\vspace{-1mm}
	\includegraphics[width=17.3cm]{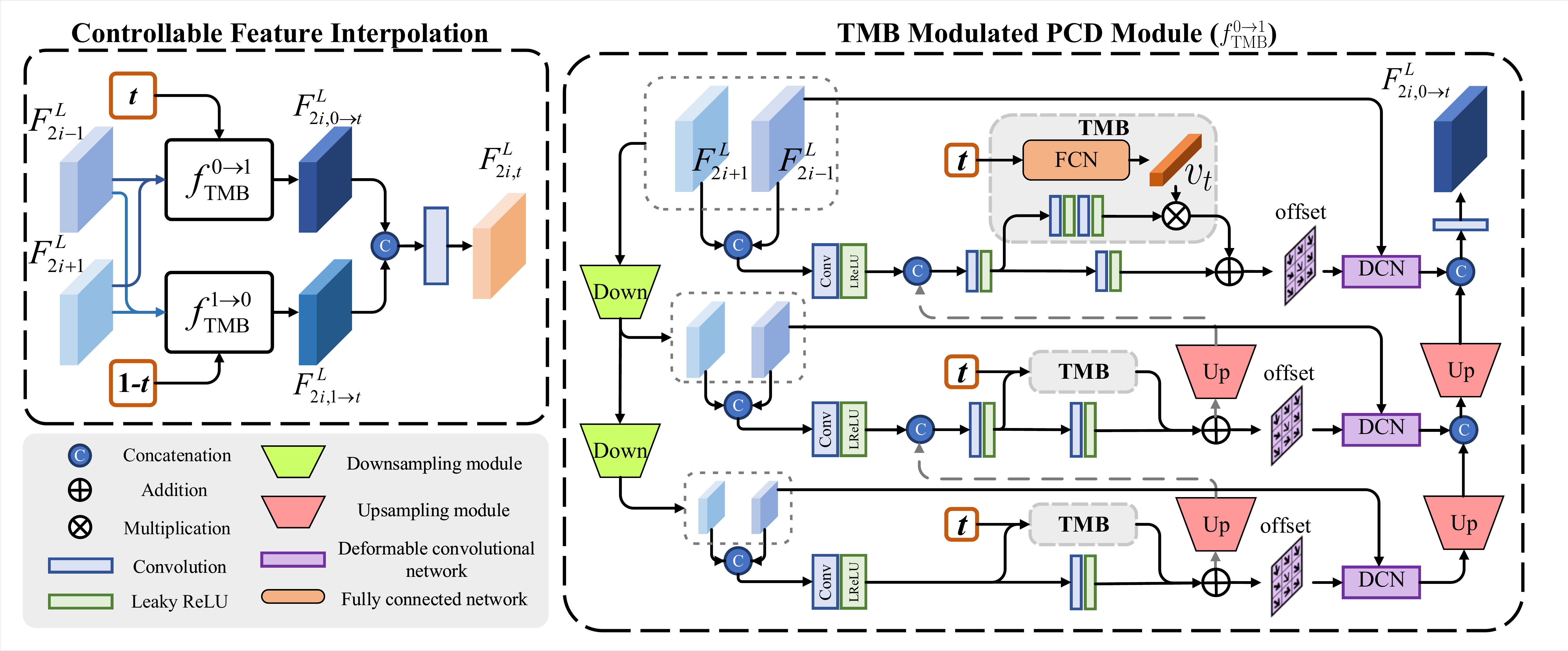}
	\vspace{-3mm}
	\caption{
	\textbf{Proposed Temporal Modulation Block (TMB)} modulated Pyramid, Cascading and Deformable (PCD) module~\cite{wang2019edvr} for controllable feature interpolation.
	$f_{\text{TMB}}^{0\rightarrow1}$ ($f_{\text{TMB}}^{1\rightarrow0}$) is the PCD module modulated by our TMB block to model the forward (backward) motion.
	Our TMB modulates all the three levels of the PCD module, by transforming the temporal hyper-parameter $t$ into a modulation vector $v_{t}$ via a fully connected network (FCN) consisted of three convolutional layers.
	}
	\label{moduleA}
	\vspace{-4mm}
 \end{figure*}

\section{Proposed Method}
\label{sec:overall}
In this section, we first overview our Temporal Modulation Network (TMNet) for STVSR in~\S\ref{sec:overview}.
Then, we introduce our Temporal Modulation Block for controllable feature interpolation in~\S\ref{sec:TMB}.
%
We present temporal feature fusion in~\S\ref{sec:LTFC}, and high-resolution 
reconstruction in \S\ref{sec:reconstruct}.
Finally, the training details are given in \S\ref{sec:details}.

\subsection{Network Overview}
\label{sec:overview}
As illustrated in~\figref{method}, our TMNet consists of three seamless stages: controllable feature interpolation, temporal feature fusion, and high-resolution reconstruction.

\noindent
\textbf{Controllable feature interpolation}.\
Given a sequence of low-frame-rate and low-resolution video $\mathcal{I}^{L}=\{\bm{I}_{2i-1}^{L}\}_{i=1}^{n}$, our TMNet firstly extracts the corresponding initial feature maps $\{\bm{F}_{2i-1}^{L}\}_{i=1}^{n}$ through 5 residual blocks.
To perform temporally controllable feature interpolation, we propose a Temporal Modulation Block (TMB) to modulate the deformable convolution kernels with a temporal hyper-parameter $t$.\
Here, $t$$\in$$(0,1)$ indicates the (arbitrary) moment at which we plan to interpolate a feature map $\bm{F}_{2i,t}^{L}$ from the feature maps $\bm{F}_{2i-1}^{L}$ and $\bm{F}_{2i+1}^{L}$ of two adjacent frames $\bm{I}_{2i-1}^{L}$ and $\bm{I}_{2i+1}^{L}$, respectively.\
Finally, we obtain a feature sequence $\mathcal{F}^{L}=\{\bm{F}_{1}^{L},\bm{F}_{2,t}^{L},\bm{F}_{3}^{L},...,\bm{F}_{2n-2,t}^{L},\bm{F}_{2n-1}^{L}\}$ of high-frame-rate and low-resolution video frames.

\noindent
\textbf{Temporal feature fusion}.
The extracted (or interpolated) feature maps in $\mathcal{F}^{L}$ are often of low-quality, since
they are extracted from individual LR frames (or interpolated by the initial feature maps of adjacent LR frames).
Thus, we propose a Locally-temporal Feature Comparison (LFC) module, to refine every feature map in $\mathcal{F}^{L}$ with the help of the feature maps of adjacent frames.
After the local feature refinement, we further improve the feature maps in $\mathcal{F}^{L}$ by performing globally-temporal feature fusion (GFF).
This is implemented by employing a Bi-directional Deformable ConvLSTM (BDConvLSTM) network~\cite{xiang2020zooming}, to consecutively aggregate the useful information from individual feature maps along the temporal direction.
Both the LFC and GFF fusion modules well exploit the intra-correlation between spatial and temporal dimensions to improve the quality of feature maps in $\mathcal{F}^{L}$.
In the end, we obtain the sequence of improved feature maps $\mathcal{F}_{GFF}^{L}$.

\noindent
\textbf{High-resolution reconstruction}.
Here, we feed the sequence of feature maps $\mathcal{F}_{GFF}^{L}$ into 40 residual blocks to improve their quality along the spatial dimension.
Next, we increase the spatial resolution of these improved feature maps via the widely used Pixel-Shuffle layers~\cite{subpixel}, and output the final high-frame-rate and high-resolution video sequence $\mathcal{I}^{H}=\{\bm{I}_{1}^{H},\bm{I}_{2,t}^{H},...,\bm{I}_{2n-2,t}^{H},\bm{I}_{2n-1}^{H}\}$.

\subsection{Controllable Feature Interpolation}
\label{sec:TMB}

Given a sequence of low-frame-rate and low-resolution video frames $\mathcal{I}^{L}=\{\bm{I}_{2i-1}^{L}\}_{i=1}^{n}$, we first extract the corresponding initial features $\mathcal{F}^{L}=\{\bm{F}_{2i-1}^{L}\}_{i=1}^{n}$ via five residual blocks.
Each residual block contains a sequence of ``Conv-ReLU-Conv'' operations with a skip connection.
For any two adjacent frames $\bm{I}_{2i-1}^{L}$ and $\bm{I}_{2i+1}^{L}$ ($i\in\{1,...,n-1\}$), our goal here is to interpolate the feature of intermediate frame at an arbitrary moment $t\in(0,1)$.
To this end, we need to estimate the motion cues from $\bm{I}_{2i-1}^{L}$ to the intermediate frame (forward) and that from $\bm{I}_{2i+1}^{L}$ to the intermediate frame (backward).
Previous STVSR methods~\cite{wang2019edvr,xiang2020zooming} utilized the Pyramid, Cascading and Deformable (PCD) module to estimate the offset between $\bm{F}_{2i-1}^{L}$ and $\bm{F}_{2i+1}^{L}$ as the motion cues, to align and interpolate the features of the intermediate frame under the deformable convolutional framework~\cite{zhu2019deformable}.
However, the vanilla PCD module could only estimate the motion to a predefined moment, which is fixed in both training and inference stages.

To overcome this limitation, we propose a Temporal Modulation Block (TMB) to modulate the learned offset between $\bm{F}_{2i-1}^{L}$ and $\bm{F}_{2i+1}^{L}$.
The modulation is controlled by a hyper-parameter $t\in(0,1)$, indicating an arbitrary moment that we plan to interpolate a new frame.
This enables our TMNet to control the feature interpolation process upon the initial feature maps $\bm{F}_{2i-1}^{L}$ and $\bm{F}_{2i+1}^{L}$ of two adjacent frames $\bm{I}_{2i-1}^{L}$ and $\bm{I}_{2i+1}^{L}$ in the input video.
The PCD module modulated by our TMB block can estimate the forward and backward motions and interpolate the feature map $\bm{F}_{2i,t}^{L}$ of a new frame at the arbitrary moment $t\in(0,1)$.

Denote by $f_{\text{TMB}}^{0\rightarrow1}$ and $f_{\text{TMB}}^{1\rightarrow0}$ the PCD modules modulated by our TMB block, to model the forward and backward motions, respectively.
Here, we perform modulated feature interpolation from the forward and backward directions:
\vspace{-2mm}
\begin{equation}
\vspace{-2mm}
\begin{aligned}
\bm{F}_{2i,0\rightarrow t}^{L}=f &_{\text{TMB}}^{0\rightarrow1}(\bm{F}_{2i-1}^{L},\bm{F}_{2i+1}^{L},t), 
\\
\bm{F}_{2i,1\rightarrow t}^{L}=f &_{\text{TMB}}^{1\rightarrow0}(\bm{F}_{2i-1}^{L},\bm{F}_{2i+1}^{L},1-t), \\
\end{aligned}
\label{f_tmb}
\vspace{-0mm}
\end{equation}
where $\bm{F}_{2i,0\rightarrow t}^{L}$ and $\bm{F}_{2i,1\rightarrow t}^{L}$ are the interpolated features aligned from the feature maps $\bm{F}_{2i-1}^{L}$ and $\bm{F}_{2i+1}^{L}$ of the adjacent frames. 
Note that the two TMB-modulated PCD modules share the same network structure but have different weights.
Here we only take the $f_{\text{TMB}}^{0\rightarrow1}$ as an example to explain how the PCD modules modulated by our TMB work on modeling the forward motion.
The PCD module $f_{\text{TMB}}^{1\rightarrow0}$ modeling the backward motion can be similarly explained.

As shown in \figref{moduleA}, the PCD module has three levels to estimate the motion in different scales.
To realize flexible modulation on the temporal dimension, we embed our TMB block into each level of the vanilla PCD module independently, to modulate the offset before the deformable convolutional network (DCN).
The benefits of adding our TMB block to all three levels of the PCD module will be verified in \S\ref{sec:experiment}.
To adaptively modulate the offset by our TMB at different levels of PCD, we use three convolutional layers to map the temporal hyper-parameter $t$ onto a modulation vector $\bm{v}_{t}$ of size $1\times1\times64$.
To well exploit the motion cues for precise modulation, we feed the features in each vanilla PCD level into two convolutional layers to enlarge their receptive fields.
Then, the generated feature is multiplied with the modulation vector $\bm{v}_{t}$ along the channel dimension, to produce the TMB-modulated features.
For robustness, we add the TMB-modulated features with the corresponding pre-modulated features before DCN.

Once obtaining the modulated feature maps $\bm{F}_{2i,0\rightarrow t}^{L}$ and $\bm{F}_{2i,1\rightarrow t}^{L}$, we interpolate the intermediate feature $\bm{F}_{2i,t}^{L}$ via channel-wise concatenation ``$[\cdot,\cdot]$'' followed by a $1\times1$ convolution layer $f_{1\times1}$ as:
\vspace{-2mm}
\begin{equation}
\vspace{-2mm}
\begin{aligned}
\bm{F}_{2i,t}^{L}= f_{1\times1}([\bm{F}_{2i,0\rightarrow t}^{L}, \bm{F}_{2i,1\rightarrow t}^{L}]).
\end{aligned}
\label{f}
\end{equation}
Now, we obtain the features of the interpolated sequence $\mathcal{F}^{L}={\{\bm{F}_{1}^{L},\bm{F}_{2,t}^{L},\bm{F}_{3}^{L},...,\bm{F}_{2n-2,t}^{L},\bm{F}_{2n-1}^{L}\}}$ for the high-frame-rate and low-resolution video.
Next, we perform feature fusion along the temporal dimension.

\subsection{Temporal Feature Fusion}
\label{sec:LTFC}

Here, the initial features are extracted (or interpolated) from individual (or adjacent) frames.
There is considerable leeway to improve their quality.
But we also feed the initial features into the Pixel-Shuffle part of our TMNet.

\begin{figure}[t]
\vspace{-1mm}
\includegraphics[width=8.3cm]{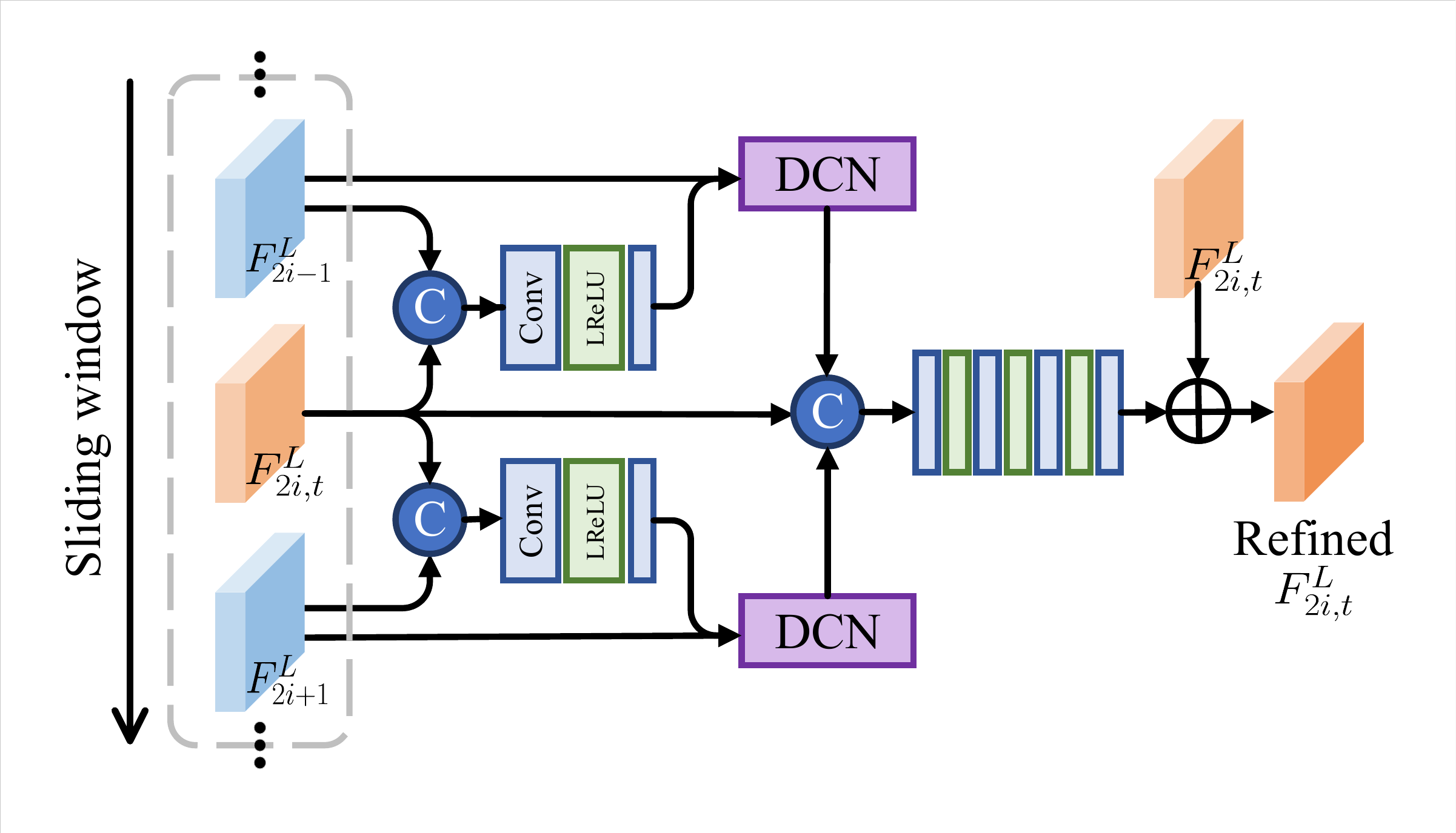}
\vspace{-7mm}
\caption{\textbf{Proposed Locally-temporal Feature Comparison} (LFC) module refines the interpolated feature $\bm{F}_{2i,t}^{L}$ by exploiting short-term motion cues among adjacent frames.}
\vspace{-4mm}
\label{moduleB}
\end{figure}

\noindent
\textbf{Locally-temporal feature comparison}. 
%
It is essential to maintain short-term temporal consistency for each current frame.
For this purpose, we propose a Locally-temporal Feature Comparison (LFC) module to exploit the complementary information (e.g., motion cues) from adjacent frames.
As illustrated in~\figref{moduleB}, to refine the feature map $\bm{F}_{2i,t}^{L}$ of current frame from adjacent feature maps $\bm{F}_{2i-1}^{L}$ and $\bm{F}_{2i+1}^{L}$,
we concatenate current frame ($\bm{F}_{2i,t}^{L}$) and adjacent frames ($\bm{F}_{2i-1}^{L}$, $\bm{F}_{2i+1}^{L}$), and employ two convolutional layers to learn the offset in the deformable convolutional framework~\cite{zhu2019deformable}.
Note that we learn two offsets to describe the motion cues in the forward (from $\bm{I}_{2i-1}^{L}$ to current frame) and the backward (from $\bm{I}_{2i+1}^{L}$ to current frame) directions.
Then, the learned offset from forward (or backward) direction is used to align the feature map $\bm{F}_{2i-1}^{L}$ of previous (or $\bm{F}_{2i+1}^{L}$ of next) frame with that of the current frame, via one deformable convolutional layer.
After the alignment, we concatenate the aligned feature maps of two adjacent frames with that of the current frame, and perform feature comparison via four $1\times1$ convolutional layers and an addition operation.
For the first (or last) frame, the previous (or next) adjacent frame is just itself. 
Now we get a refined feature sequence $\mathcal{F}_{LFC}^{L}$.

\noindent
\textbf{Globally-temporal feature fusion}.
The feature sequence refined by our LFC module is able to maintain short-term consistency in the interpolated video.
But it would fail on large or fast motions, since LFC lacks the capability of modeling the motions over the whole video. 
To tackle this problem, we propose to exploit the long-term information in videos by globally-temporal feature fusion.
Inspired by~\cite{xiang2020zooming}, we feed the feature sequence $\mathcal{F}_{LFC}^{L}$ generated by our LFC into the BDConvLSTM network, and obtain the features $\mathcal{F}_{GFF}^{L}$ in long-term temporal consistency.

As will be illustrated in the experimental section, our short-term LFC module and the long-term BDConvLSTM indeed boost the performance of our TMNet on STVSR.
 
\subsection{High-Resolution Reconstruction}
\label{sec:reconstruct}
Until now, the intra-correlation of temporal and spatial dimensions is well explored to obtain the high-quality feature sequence $\mathcal{F}_{GFF}^{L}$ of the whole video.
Then, we perform spatial refinement for the feature maps via 40 residual blocks, and get the refined feature maps $\mathcal{F}^{H}$.
Then we add the features $\mathcal{F}^{H}$ with the corresponding initial feature maps in $\mathcal{F}^{L}$, and obtain the reconstructed feature maps $\mathcal{F}_{final}^{H}$.
Finally, we feed the reconstructed feature maps $\mathcal{F}_{final}^{H}$ into two Pixel-Shuffle layers, followed by a sequence of ``Conv-LeakyReLU-Conv'' operations, to output the reconstructed HR video frames $\mathcal{I}^{H}=\{\bm{I}_{1}^{H},\bm{I}_{2,t}^{H},...,\bm{I}_{2n-2,t}^{H},\bm{I}_{2n-1}^{H}\}$.

\begin{table*}[t]
\scriptsize
\vspace{-1mm}
\centering 
\caption{
\textbf{
Comparison of PSNR, SSIM~\cite{ssim}, speed}
(\textbf{in} \texttt{fps})\textbf{, and parameters} (\textbf{in} \texttt{million})
\textbf{by different STVSR methods
}
on Vid4~\cite{su2017deep}, Vimeo-\texttt{Fast}, Vimeo-\texttt{Medium}, Vimeo-\texttt{Slow}~\cite{xue2019video}.
%
``$\uparrow$'' means that larger is better.
The speed is evaluated on Vid4~\cite{su2017deep}.
The best, second best and third best results are highlighted in \redbf{red}, \bluebf{blue} and \textbf{bold}, respectively.
\vspace{-3mm}
}
\resizebox{1.0\linewidth}{!}{
\renewcommand\arraystretch{1.3} 
		\begin{tabular}{l||cc|cc|cc|cc|c|c}
			\Xhline{1pt}
			\rowcolor[rgb]{ .9,  .9,  .9} \multicolumn{1}{c||}{Method} & \multicolumn{2}{c|}{Vid4~\cite{su2017deep}} & \multicolumn{2}{c|}{Vimeo-\texttt{Fast}} & \multicolumn{2}{c|}{Vimeo-\texttt{Medium}} & \multicolumn{2}{c|}{Vimeo-\texttt{Slow}} & Speed & Parameters
			\\
			\rowcolor[rgb]{.9, .9, .9} 
			\multicolumn{1}{c||}{VFI+(V)SR / STVSR} & PSNR$\uparrow$ & SSIM$\uparrow$ & PSNR$\uparrow$ & SSIM$\uparrow$ & PSNR$\uparrow$ & SSIM$\uparrow$ & PSNR$\uparrow$ & SSIM$\uparrow$ & \texttt{fps}$\uparrow$ & \texttt{million}$\downarrow$ 
			\\
			\hline
			SuperSloMo~\cite{jiang2018super} + Bicubic & 22.84 & 0.5772 & 31.88 & 0.8793 & 29.94 & 0.8477 & 28.37 & 0.8102 & - & \textbf{19.8}
			\\
			\rowcolor[rgb]{ .95,  .95,  .95}
			SuperSloMo~\cite{jiang2018super} + RCAN~\cite{rcan}  & 23.80 & 0.6397 & 34.52 & 0.9076 & 32.50 & 0.8884 & 30.69 & 0.8624 & 2.49 & 19.8+16.0
			\\
			SuperSloMo~\cite{jiang2018super} + RBPN~\cite{haris2019recurrent}  & 23.76 & 0.6362 & 34.73 & 0.9108 & 32.79 & 0.8930 & 30.48 & 0.8584 & 2.06 & 19.8+12.7
			\\
			\rowcolor[rgb]{ .95,  .95,  .95}
			SuperSloMo~\cite{jiang2018super} + EDVR~\cite{wang2019edvr}  & 24.40 & 0.6706 & 35.05 & 0.9136 & 33.85 & 0.8967 & 30.99 & 0.8673 & 6.85 & 19.8+20.7
			\\
			\hline
			SepConv~\cite{niklaus2017video} + Bicubic & 23.51 & 0.6273 & 32.27 & 0.8890 & 30.61 & 0.8633 & 29.04 & 0.8290 & - & 21.7
			\\
			\rowcolor[rgb]{ .95,  .95,  .95}
			SepConv~\cite{niklaus2017video} + RCAN~\cite{rcan} & 24.92 & 0.7236 & 34.97 & 0.9195 & 33.59 & 0.9125 & 32.13 & 0.8967 & 2.42 & 21.7+16.0
			\\
			SepConv~\cite{niklaus2017video} + RBPN~\cite{haris2019recurrent} & 26.08 & 0.7751 & 35.07 & 0.9238 & 34.09 & 0.9229 & 32.77 & 0.9090 & 2.01 & 21.7+12.7
			\\
			\rowcolor[rgb]{ .95,  .95,  .95}
			SepConv~\cite{niklaus2017video} + EDVR~\cite{wang2019edvr} & 25.93 & 0.7792 & 35.23 & 0.9252 & 34.22 & 0.9240 & 32.96 & 0.9112 & 6.36 & 21.7+20.7
			\\
			\hline
			DAIN~\cite{bao2019depth} + Bicubic & 23.55 & 0.6268 & 32.41 & 0.8910 & 30.67 & 0.8636 & 29.06 & 0.8289 & - & 24.0
			\\
			\rowcolor[rgb]{ .95,  .95,  .95}
			DAIN~\cite{bao2019depth} + RCAN~\cite{rcan} & 25.03 & 0.7261 & 35.27 & 0.9242 & 33.82 & 0.9146 & 32.26 & 0.8974 & 2.23 & 24.0+16.0
			\\
			DAIN~\cite{bao2019depth} + RBPN~\cite{haris2019recurrent} & 25.96 & 0.7784 & 35.55 & 0.9300 & 34.45 & 0.9262 & 32.92 & 0.9097 & 1.88 & 24.0+12.7
			\\
			\rowcolor[rgb]{ .95,  .95,  .95}
			DAIN~\cite{bao2019depth} + EDVR~\cite{wang2019edvr} & \textbf{26.12} & 0.7836 & 35.81 & 0.9323 & 34.66 & 0.9281 & \textbf{33.11} & 0.9119 & 5.20 & 24.0+20.7
			\\
			\hline
			STARnet~\cite{haris2020space} & 26.06 & \redbf{0.8046}  & \textbf{36.19} & \textbf{0.9368} & \textbf{34.86} & \textbf{0.9356} & 33.10 & \redbf{0.9164} & \textbf{14.08} & 111.61
			\\
			\rowcolor[rgb]{ .95,  .95,  .95}
			Zooming Slow-Mo~\cite{xiang2020zooming} & \bluebf{26.31} & \textbf{0.7976} & \bluebf{36.81} & \bluebf{0.9415} & \bluebf{35.41} & \bluebf{0.9361} & \bluebf{33.36} & \textbf{0.9138} & \redbf{16.50} & \redbf{11.10}
			\\
			\textbf{TMNet (Ours)} & \redbf{26.43}  & \bluebf{0.8016} & \redbf{37.04}  & \redbf{0.9435}  & \redbf{35.60} & \redbf{0.9380} & \redbf{33.51} & \bluebf{0.9159} & \bluebf{14.69} & \bluebf{12.26}
			\\
			\hline
		\end{tabular}
	}
	\vspace{-4mm}
	\label{tab:sota}
\end{table*}

\subsection{Training Details}
\label{sec:details}

\noindent
\textbf{Implementation details}.\
We employ the Adam optimizer~\cite{adam} with $\beta_{1} = 0.9$ and $\beta_{2} = 0.999$ to optimize our TMNet with the Charbonnier loss function~\cite{lai2017deep}, as suggested in~\cite{xiang2020zooming}.
The learning rate is initialized as $4\times10^{-4}$, and is decayed to $1\times10^{-7}$ with a cosine annealing~\cite{loshchilov2016sgdr} for every 150,000 iterations.
We initialize the parameters of our TMNet by Kaiming initialization~\cite{he2015delving} without pre-trained weights.
The batch size is 24.
Our TMNet, implemented in PyTorch~\cite{pytorch} and Jittor~\cite{hu2020jittor}, is trained in a total of 600,000 iterations on four RTX 2080Ti GPUs, which takes about 8.71 days (209.04 hours).
For each input video clip, we randomly crop it into a sequence of downsampled patches of size $32\times32$.
For data argumentation, we horizontal-flip each frame, and randomly rotate it with $90^{\circ}$, $180^{\circ}$, or $270^{\circ}$.

\noindent
\textbf{Network training}.\
When directly trained with the proposed TMB block, our TMNet suffers from clear performance drops on STVSR, as shown in our experiments.
One possible reason is that our TMNet can not accurately estimate the motion cues to interpolate an intermediate frame at the arbitrary moment $t\in(0,1)$ since our TMB does not know the moment before the training modulated feature.
To resolve this problem, we propose to train our TMNet by a two-step strategy: \texttt{Step 1}, we train our main TMNet without the proposed TMB block;
\texttt{Step 2}, we only train our TMB block while fixing the trained main network.

In \texttt{Step 1}, we train our TMNet on the Vimeo-90K dataset~\cite{xue2019video}, which will be introduce in~\S\ref{sec:setup}.
%
This dataset consists of 7-frame video clips.
For each clip, the 1-st, 3-rd, 5-th, and 7-th LR frames are input into our TMNet as low-frame-rate and low-resolution video.
We set $t=0.5$ to get rid of the TMB block from our TMNet, and learn to generate the 7-frame high-resolution and high-frame-rate video.
This enables our TMNet to fairly compare with previous STVSR methods~\cite{su2017deep,wang2019edvr,haris2020space,xiang2020zooming}.
For supervision, we calculate the loss function over the corresponding 7-frame HR video clip in the Vimeo-90K dataset~\cite{xue2019video}.

%
In \texttt{Step 2}, we fix the learned weights of our main network, and only train our TMB block for temporal modulation.\ The training is performed on the Adobe240fps dataset~\cite{su2017deep}, which is in high-frame-rate and suitable for training our TMB block.\ We also split it into groups of 7-frame video clips.\ For each clip, the 1-st and 7-th HR frames are downsampled as the inputs of our TMNet.\ We set the temporal hyper-parameter $t$$\in$$\{\frac{1}{6},\frac{2}{6},\frac{3}{6},\frac{4}{6},\frac{5}{6}\}$ to interpolate 5 intermediate frames.
This step costs 35.26 minutes.

\section{Experiments}
\label{sec:experiment}

\subsection{Experimental Setup}
\label{sec:setup}
\begin{figure*}[t]
\vspace{-1mm}
	\centering
	\includegraphics[width=\linewidth]{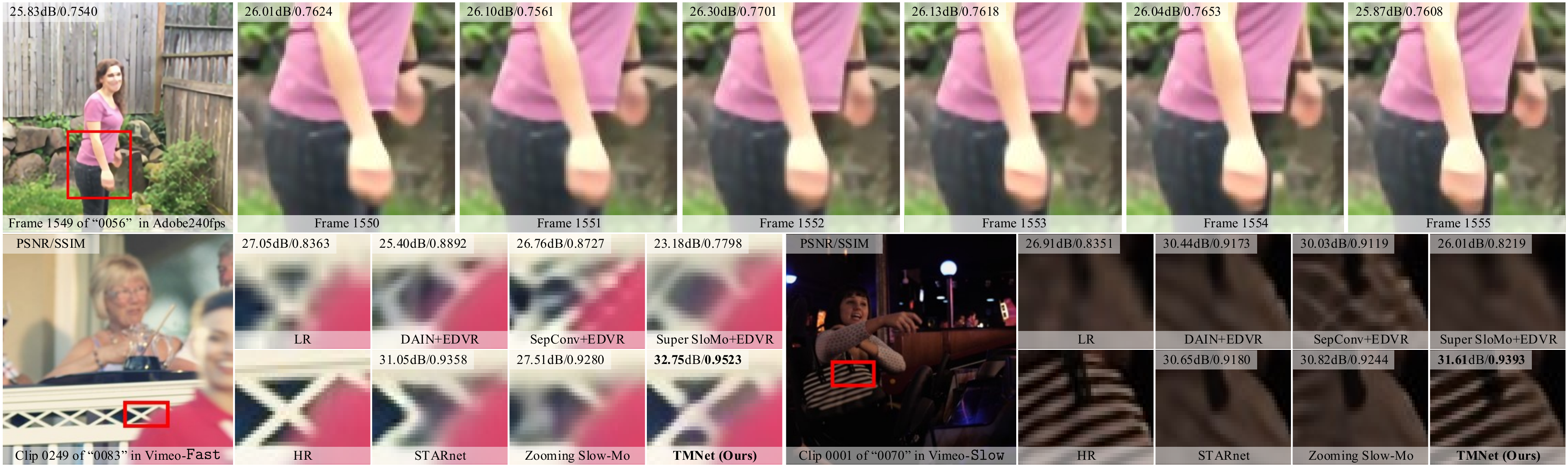}
	\vspace{-7mm}
	\caption{\textbf{Qualitative and quantitative results of different methods on STVSR}.
	The test video clips are from the Adobe240fps~\cite{su2017deep} (1-st row), Vimeo-\texttt{Fast}~\cite{xue2019video} (2-nd row, left) and Vimeo-\texttt{Slow}~\cite{xue2019video} (2-nd row, right) datasets.}
	\vspace{-5mm}
	\label{fig:visualization}
\end{figure*}

\noindent
\textbf{Dataset}.
We use Vimeo-90K \texttt{septuplet} dataset~\cite{xue2019video} as the training set.
It contains 91,701 video sequences, extracted from 39K video clips selected from Vimeo-90K.
Each sequence contains 7 continuous frames of resolution $448\times256$.
The Vid4~\cite{vid4} and Vimeo-90K \texttt{test} set are used as evaluation datasets.
As suggested in~\cite{xiang2020zooming}, we split the Vimeo-90K \texttt{septuplet test} set into three subsets of \texttt{Fast} motion, \texttt{Medium} motion, and \texttt{Slow} motion, which include 1225, 4977, and 1613 video clips, respectively.
We also remove 5 video clips from the original \texttt{Medium} motion set and 3 clips from the \texttt{Slow} motion set, which contain only all-black backgrounds.

To make our TMNet feasible for controllable feature interpolation, we train our TMB block individually on the Adobe240fps dataset~\cite{su2017deep}.
It has 133 videos (in 720P) taken with hand-held cameras, and is randomly split into the \texttt{train}, \texttt{val}, and \texttt{test} subsets with 100, 16, and 17 videos, respectively.
For each video, we split it into groups of 7-frame video clips.
We feed the 1-st and 7-th frames in each clip into our TMNet to generate 5 intermediate frames.

We downsample the HR frames to create the LR frames by bicubic interpolation, with a factor of 4.

\noindent
\textbf{Evaluation metric}.\
We employ the widely used Peak Signal-to-Noise Ratio (PSNR) and Structural Similarity Index (SSIM)~\cite{ssim} to evaluate different methods on the STVSR task.
The PSNR and SSIM metrics are calculated on the Y channel of the YCbCr color space, as favored by previous VSR~\cite{haris2019recurrent,wang2019edvr} and STVSR~\cite{xiang2020zooming} methods.

\subsection{Comparison to State-of-the-arts}

\noindent
\textbf{Comparison methods}.
We compare our TMNet with state-of-the-art two-stage and one-stage STVSR methods.
For the two-stage STVSR methods, we perform video frame interpolation (VFI) by SuperSloMo~\cite{jiang2018super}, DAIN~\cite{bao2019depth} or SepConv~\cite{niklaus2017video}, and perform video super-resolution (VSR) by Bicubic Interpolation (BI), RCAN~\cite{rcan}, RBPN~\cite{haris2019recurrent} or EDVR~\cite{wang2019edvr}.
For one-stage STVSR methods, we compare our TMNet with the recently developed Zooming SlowMo~\cite{xiang2020zooming} and STARnet~\cite{haris2020space}.
To fairly compare with these competitors, we set $t=0.5$ in our TMNet to generate the frame at the middle moment of any two adjacent frames.
That is, the 1-st, 3-rd, 5-th, and 7-th LR frames of each clip in Vimeo-90K are fed into our TMNet to reconstruct the 7 HR frames.
All these methods are trained on the Vimeo-90K \texttt{septuplet} dataset~\cite{xue2019video}, evaluated on the Vimeo-90K \texttt{test} set~\cite{xue2019video} and the Vid4~\cite{su2017deep} dataset.

\noindent
\textbf{Objective results}.
We list the quantitative comparison results in Table~\ref{tab:sota}.
As suggested in~\cite{xiang2020zooming}, we omit the baseline models with Bicubic Interpolation when comparing the speed.
One can see that our TMNet outperforms the Zooming SlowMo~\cite{xiang2020zooming} by 0.12dB, 0.23dB, 0.19dB, and 0.15dB on the Vid4, Vimeo-\texttt{Fast}, Vimeo-\texttt{Medium}, and Vimeo-\texttt{Slow} datasets in terms of PSNR.
On SSIM~\cite{ssim}, our TMNet achieves better results than the competitors in most cases, but is only slightly inferior to STARnet~\cite{haris2020space} on Vid4~\cite{vid4} and Vimeo-\texttt{Slow}.
However, our TMNet needs only one-ninth of the parameters in STARnet.
On speed, one-stage methods~\cite{haris2020space,xiang2020zooming} run much faster than two-stage ones~\cite{niklaus2017video,jiang2018super,bao2019depth,haris2019recurrent,wang2019edvr}.
Our TMNet runs at 14.69fps, and is only slower than Zooming Slow-Mo~\cite{xiang2020zooming}.
All these results validate the effectiveness of our TMNet on STVSR.

\noindent
\textbf{Visualization}.
In the 1-st row of Figure~\ref{fig:visualization}, we present the 5 intermediate frames (Frame 1550 to Frame 1554) interpolated by our TMNet on the sequence ``0056'' from the Adobe240fps \texttt{test} set~\cite{su2017deep}, given the Frame 1549 and Frame 1555 as inputs.
It can be seen that our TMNet is able to perform flexible frame interpolation for STVSR.
In the 2-nd row of Figure~\ref{fig:visualization}, we show the reconstructed frames by different STVSR methods from Vimeo-\texttt{Fast} and and Vimeo-\texttt{Slow} datasets~\cite{xue2019video} generated by the competing methods respectively.
We observe that our TMNet, with the proposed LFC module, can restore more clearly the structures and textures than the competitors.
For example, on the Clip ``0001'' in the sequence ``0070'' of Vimeo-\texttt{Slow} datasets, our TMNet reconstructs clearly the texture pattern on the bag.
In summary, our TMNet demonstrates flexible and powerful STVSR ability quantitatively and qualitatively.
%
More visual comparison on the Vid4~\cite{vid4}, Vimeo-90K \texttt{test} set~\cite{su2017deep}, and Adobe240fps~\cite{xue2019video} datasets are provided in the \textsl{Supplementary File}, because of page limitation.



\subsection{Ablation Study}

Here, we conduct detailed examinations of our TMNet on STVSR.
Specifically, we assess
1) the importance of our Temporal Modulation Block (TMB) for controllable feature interpolation;
2) different strategies that our TMB block modulates the PCD module;
3) how to design our TMB block;
4) how our Locally-temporal Feature Comparison (LFC) module contribute to the temporal feature fusion in our TMNet;
%
5) the combination of high-quality feature maps $\mathcal{F}^{H}$ and initial feature maps $\mathcal{F}^{L}$ for STVSR.

\noindent
\textbf{1. Does our TMB block contribute to controllable feature interpolation}?
To answer this question, we compare our TMNet with previous STVSR methods~\cite{xiang2020zooming,haris2020space} on generating intermediate frames from two adjacent frames.
Due to limited space, we provide the comparison of  visual results on Adobe240fps \texttt{test} set~\cite{su2017deep} in the \textsl{Supplementary File}.
We observe that our TMNet with TMB block indeed exhibits temporally controllable STVSR performance.

\noindent
\textbf{2. How different strategies that our TMB block modulate the PCD module influence our TMNet on STVSR}?
The PCD module~\cite{wang2019edvr} has a three-level pyramid structure: the 1-st level \texttt{L1}; the 2-nd level \texttt{L2} is downsampled from the features in \texttt{L1} by convolution filters at a stride of 2; similarly, the 3-rd level \texttt{L3} is downsampled from \texttt{L2} by a stride of 2.
%
In our TMNet, the proposed TMB modulates all the three levels of the PCD module.
But our TMB can also modulate only one level (\texttt{L1}, \texttt{L2}, or \texttt{L3}) of PCD, resulting three variants of our TMNet called TMB-\texttt{L1}, TMB-\texttt{L2}, and TMB-\texttt{L3}.
These variants are trained on the Adobe240fps \texttt{train} set~\cite{su2017deep} and evaluate them on \texttt{test} set.
As shown in Table~\ref{tab:tmbpcd}, the three variants perform in descending order, indicating that the 1-st level of PCD is more important for temporal modulation.
By modulating all three levels of PCD, our TMNet outperforms the three variants on STVSR, by better exploiting the motion cues of videos.
\begin{table}[h]
\vspace{-2.5mm}
\scriptsize
\centering
\caption{\textbf{PSNR results on Adobe240fps \texttt{test} set by different strategies that our TMB modulates the PCD.}}
\vspace{-3mm}
	\resizebox{\columnwidth}{!}{
		\begin{tabular}{l||cccc}
			\Xhline{1pt}
			\rowcolor[rgb]{ .9,  .9,  .9}
			\multicolumn{1}{c||}{\multirow{-1}{*}{Variant}} & TMB-\texttt{L1} & TMB-\texttt{L2} & TMB-\texttt{L3} & TMNet \\
			\hline
			PSNR (dB) & 26.92 & 26.82 & 26.60 & \textbf{26.95} \\
			\hline
		\end{tabular}
	}
	\label{tab:tmbpcd}
\vspace{-3mm}
\end{table}

\noindent
\textbf{3. How to design our TMB block}?
The goal of our TMB is to transform the hyper-parameter $t$ into a modulation vector $v_{t}$ comfortable with the PCD module.
A trivial design of our TMB is a linear convolutional layer.
We call it TMB-\texttt{Linear}.
We train our TMB and the TMB-\texttt{Linear} on Adobe240fps \texttt{train} set~\cite{su2017deep} and evaluate them on \texttt{test} set.
The PSNR results are listd in Table~\ref{tab:designTMB}, in which the TMB-\texttt{Linear} is 0.02dB lower than our TMB with three nonlinear convolutional layers.
This shows that nonlinear transformation is only a little better than the linear one.
\begin{table}[h]
\vspace{-2.5mm}
\tiny
\centering 
\caption{\textbf{PSNR results on Adobe240fps \texttt{test} set by our TMB with linear or nonlinear design}.}
\vspace{-3mm}
	\resizebox{\columnwidth}{!}{
		\begin{tabular}{l||cc}
			\Xhline{1pt}
			\rowcolor[rgb]{ .9,  .9,  .9}
			\multicolumn{1}{c||}{\multirow{-1}{*}{Variant}} &  TMB-\texttt{Linear}  & TMB \\
			\hline
			PSNR (dB) & 26.93 & 26.95 \\
			\hline
		\end{tabular}
	}
\label{tab:designTMB}
\vspace{-3mm}
\end{table}

\noindent
\textbf{4. How important is the proposed LFC module to our TMNet}?\
Our TMNet performs a two-stage temporal feature fusion: first local fusion via LFC and then global fusion via GFF.
Thus, our TMNet can be called ``\texttt{LFC$\rightarrow$GFF}''.
Inverting the order, i.e., \texttt{GFF$\rightarrow$LFC}, makes our TMNet collapse during the training.
The main reason is that performing GFF before LFC brings noisy long-term information, confusing the learning of deformable convolution in LFC.
Thus, we do not evaluate this variant.
To study how our LFC contributes to the two-stage fusion in our TMNet, we remove LFC from our TMNet and call this variant ``\texttt{GFF}''.
Besides, the features from LFC and GFF can be concatenated and fused by a convolutional layer, resulting in a variant ``\texttt{LFC}+\texttt{GFF}''.
We train our TMNet and its variants on the Vimeo-90K \texttt{septuplet} dataset and evaluate them on the Vid4~\cite{vid4}, Vimeo-\texttt{Fast}, Vimeo-\texttt{Medium}, and Vimeo-\texttt{Slow} datasets.
The PSNR results are listed in Table~\ref{tab:lfc}.
One can see that our TMNet (\texttt{LFC$\rightarrow$GFF}) achieves the best results on all cases, and outperforms \texttt{GFF} by 0.07dB on Vid4, 0.17dB on Vimeo-\texttt{Fast}, 0.15dB on Vimeo-\texttt{Medium}, and 0.11dB on Vimeo-\texttt{Slow}.
This indicates that our LFC module is essential to the success of our TMNet on STVSR, by exploiting short-term motion cues among adjacent frames.
\begin{table}[h]
\vspace{-2.5mm}
\caption{\textbf{Comparison of PSNR (dB) results by different variants of our TMNet} on STVSR datasets.}
\vspace{-3mm}
	\resizebox{\columnwidth}{!}{
		\begin{tabular}{l||ccc}
			\Xhline{1pt}
			\rowcolor[rgb]{ .9,  .9,  .9}
			\multicolumn{1}{c||}{\multirow{-1}{*}{Variant}} & \texttt{GFF} & \texttt{LFC}+\texttt{GFF} & \texttt{LFC$\rightarrow$GFF}\\
			\hline
			Vid4~\cite{vid4} & 26.36 & 26.35 & \textbf{26.43} \\
			\rowcolor[rgb]{ .95,  .95,  .95}
			Vimeo-\texttt{Fast} & 36.87 & 36.90 & \textbf{37.04} \\
			Vimeo-\texttt{Medium} & 35.45 & 35.47 & \textbf{35.60} \\
			\rowcolor[rgb]{ .95,  .95,  .95}
			Vimeo-\texttt{Slow} & 33.40 & 33.43 & \textbf{33.51} \\
			\hline
		\end{tabular}
	}
\label{tab:lfc}
\vspace{-3mm}
\end{table}

\noindent
\textbf{5. The benefits of combining the high-quality feature maps $\mathcal{F}^{H}$ and the initial feature maps $\mathcal{F}^{L}$ for STVSR}.
In our TMNet, we combine the high-quality features $\mathcal{F}^{H}$ with the initial features $\mathcal{F}^{L}$ before the Pixel-Shuffle layers for final STVSR.
Since the initial features $\mathcal{F}^{L}$ largely influence our LFC module, we remove them both from our TMNet and obtain a variant ``Baseline''.
Then we add $\mathcal{F}^{L}$ to the ``Baseline'', and obtain a variant model ``+$\mathcal{F}^{L}$''.
We train our TMNet and the two variants on the Vimeo-90K \texttt{septuplet} dataset~\cite{xue2019video}, and evaluate them on Vimeo-90K \texttt{test} and Vid4~\cite{vid4} datasets.
As shown in Table~\ref{tab:skip}, the variant ``+$\mathcal{F}^{L}$'' clearly exceeds the ``Baseline''.
%
This validates that combining high-quality features $\mathcal{F}^{H}$ with initial ones $\mathcal{F}^{L}$ is helpful to our TMNet on STVSR.
\begin{table}[h]
\vspace{-2.5mm}
\tiny
\caption{\textbf{Comparison of PSNR (dB) by our TMNet and its variants on different STVSR datasets}.
}
\vspace{-3mm}
\resizebox{\columnwidth}{!}{
		\begin{tabular}{l||ccc}
			\Xhline{1pt}
			\rowcolor[rgb]{ .9,  .9,  .9}
			\multicolumn{1}{c||}{\multirow{-1}{*}{Variant}} & Baseline & +$\mathcal{F}^{L}$ & TMNet \\
			\hline
			Vid4~\cite{vid4} & 26.33 & 26.36 & \textbf{26.43} \\
			\rowcolor[rgb]{ .95,  .95,  .95}
			Vimeo-\texttt{Fast} & 36.75 & 36.87 & \textbf{37.04} \\
			Vimeo-\texttt{Medium} &  35.35 & 35.45 & \textbf{35.60} \\
			\rowcolor[rgb]{ .95,  .95,  .95}
			Vimeo-\texttt{Slow} & 33.28 & 33.40 & \textbf{33.51} \\
			\hline
		\end{tabular}
	}
\label{tab:skip}
\vspace{-4mm}
\end{table}

\section{Conclusion}
In this work, we proposed a Temporal Modulation Network (TMNet) to flexibly interpolate intermediate frames for space-time video super-resolution (STVSR).
Specifically, we introduced a Temporal Modulation Block to modulate the learning of the deformable convolution framework for controllable feature interpolation.
To well exploit motion cues, we performed short-term and long-term temporal feature fusion consisting of our proposed Locally-temporal Feature Comparison (LFC) module and a Bi-directional Deformable ConvLSTM, respectively.
Experiments on three benchmarks demonstrated the flexibility of our TMNet on interpolating intermediate frames, quantitative and qualitative advantages of our TMNet over previous methods, and effectiveness of our LFC module, for STVSR.


\section*{Author Contributions}
J.X. conceived and managed the project, contributed major ideas, designed experimental settings, and rewrote the paper. G.X. contributed ideas, implemented networks, performed experiments, and wrote the first draft. Z.L. plotted some initial figures. L.W., X.S. and M.C. contributed discussions.

{\small
\bibliographystyle{ieee_fullname}
\bibliography{tmnet}
}

\twocolumn

\begin{appendices}
\renewcommand{\thesection}{\arabic{section}}
\section{Content}
In this supplemental file, we provide 
more details of our Temporal Modulation Network (TMNet) for Space-Time Video Super-Resolution (STVSR).
Specifically, we provide
\begin{itemize}
\item the detailed network structure of our TMNet in~\S\ref{sec:architecture};
\item more details of our two-step training scheme in~\S\ref{sec:details};
\item flexibility of our TMNet on interpolating arbitrary number of intermediate frames in~\S\ref{sec:flexible};
\item more visual comparisons of our TMNet with previous STVSR methods in~\S\ref{sec:comparison};
\item how the one-stage training (instead of two-stage) influences our TMNet with TMB on STVSR in~\S\ref{sec:one-stage}.
\end{itemize}

\section{Detailed Network Structure of Our TMNet}
\label{sec:architecture}
\begin{figure*}[t]
	\centering
	\includegraphics[width=\linewidth]{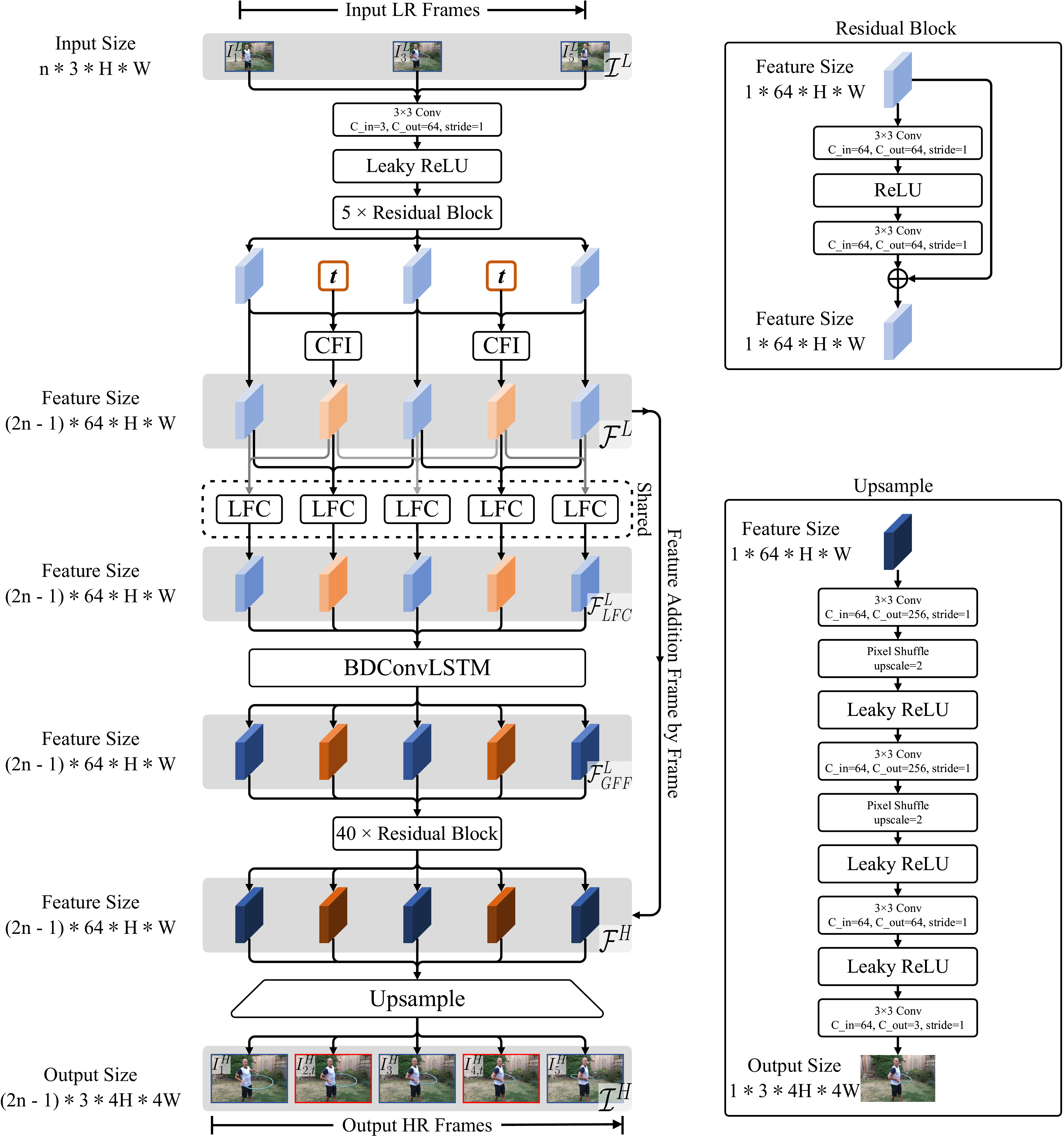}
	\vspace{-6mm}
	\caption{\textbf{Main structure of our TMNet}. 
	The basic part of ``Residual Block'' and ``Upsample'' are illustrated on the right side.
	$n$ is the number of input frames. 
	$H$ and $W$ denote the height and width of the image or feature map.
	C$\_$in and C$\_$out denote the number of input and output channels, respectively. 
	}
	\vspace{-3mm}
	\label{fig:structure}
\end{figure*}

Here, we illustrate the detailed network architecture of our proposed TMNet in Figure~\ref{fig:structure}.

\begin{figure*}[t]
	\centering
	\includegraphics[width=\linewidth]{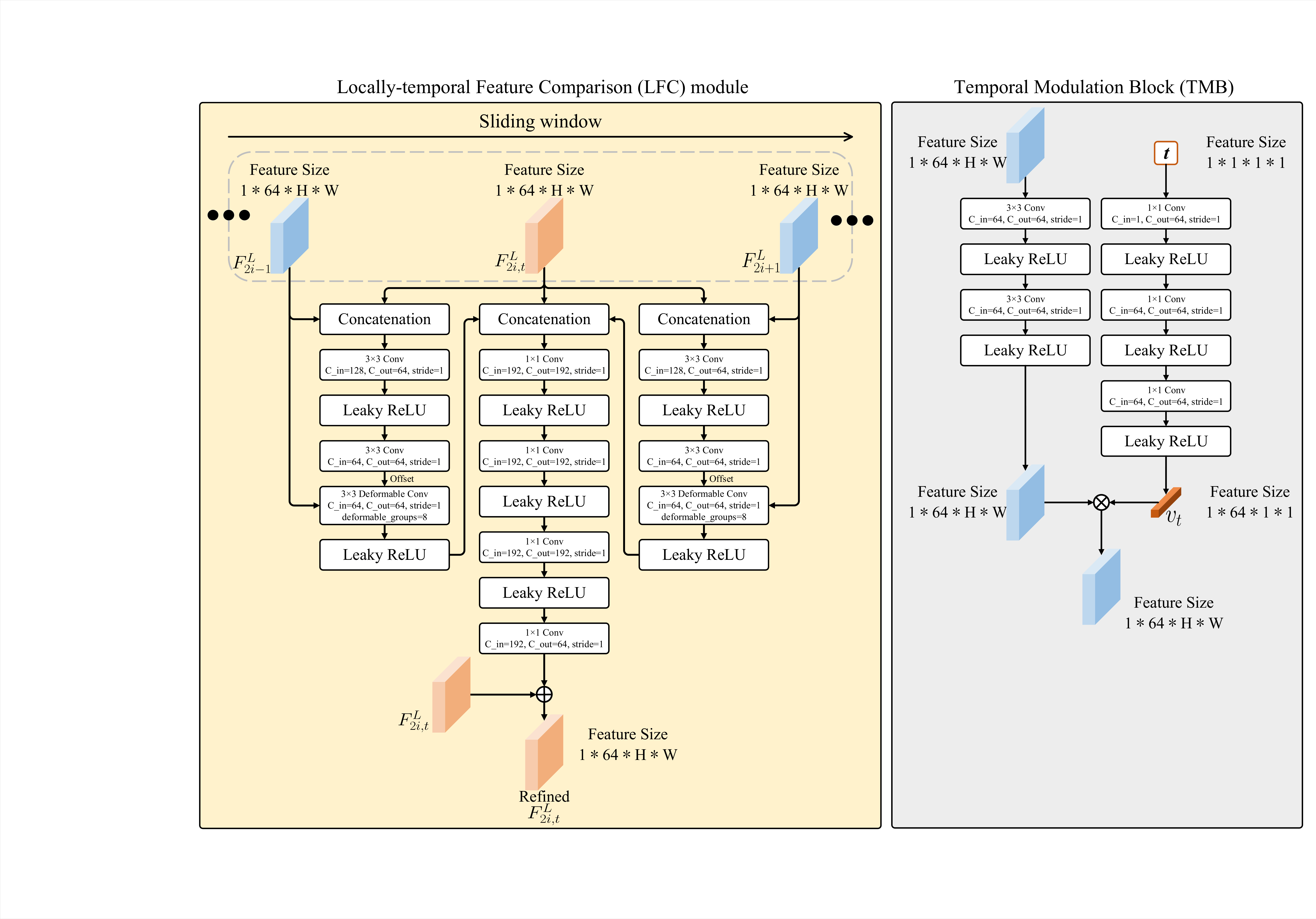}
	\vspace{-6mm}
	\caption{\textbf{Detailed structures of our Locally-temporal Feature Comparison (LFC) module (left) and Temporal Modulation Block (TMB) (right)}.
	$2i-1$, $2i$, and $2i+1$ are the indexes of frames. 
	$H$ and $W$ denote the height and width of the image or feature map.
	C$\_$in and C$\_$out denote the number of input and output channels, respectively. 
	}
	\vspace{-3mm}
	\label{fig:lfc+tmb}
\end{figure*}

We first extract the corresponding initial features $\mathcal{F}^{L}=\{\bm{F}_{2i-1}^{L}\}_{i=1}^{n}$ via five residual blocks.
Each residual block contains a sequence of ``Conv-ReLU-Conv'' operations with a skip connection.
The Controllable Feature Interpolation (CFI) is performed by the Pyramid, Cascading and Deformable (PCD) module~\cite{wang2019edvr} modulated by our proposed Temporal Modulation Block (TMB), which is illustrated in Figure 2 of our main paper.
The detailed structure of our TMB block is shown in~\ref{fig:lfc+tmb} (right).
The proposed Locally-temporal Feature Comparison (LFC) module is presented in Figure~\ref{fig:lfc+tmb} (left). 
The BDConvLSTM part is directly implemented by employing the Bi-directional Deformable ConvLSTM network in~\cite{xiang2020zooming}.
The Upsampling part contains operations of two ``Convolutions (Conv), Pixel-Shuffle, and LeakyReLU'', and one ``Conv-LeakyReLU-Conv''.

\section{More Details of Two-step Training Scheme}
\label{sec:details}

Here, we provide more details of the two-step training strategy for our TMNet.

In \texttt{Step 1}, we use the Vimeo-90K \texttt{septuplet} dataset~\cite{xue2019video} as the training set, and the Vid4~\cite{vid4}, Vimeo-\texttt{Fast}, Vimeo-\texttt{Medium}, and Vimeo-\texttt{Slow} sets as the evaluation sets.
The Vimeo-90K \texttt{septuplet}, Vimeo-\texttt{Fast}, Vimeo-\texttt{Medium}, and Vimeo-\texttt{Slow} datasets~\cite{xue2019video} consist of 7-frame video sequences, and the Vid4~\cite{vid4} dataset contains 4 video clips, which contains 41, 34, 49 and 47 frames, respectively.
All the frames in the Vid4 dataset~\cite{vid4} are split into sequences containing 7 continuous frames.
We downsample all the original HR frames to obtain the low-resolution (LR) input frames via Bicubic interpolation, by a factor of 4.
When we train our TMNet, we initialize the parameters of our TMNet by Kaiming initialization~\cite{he2015delving} without pre-trained weights.
We set $t=0.5$ to get rid of the TMB block and take the 1-st, 3-rd, 5-th, and 7-th LR frames of every sequence as a low-frame-rate and low-resolution input video to train our TMNet. 
Thus, with the supervision of the corresponding 7-frame HR video sequences in the Vimeo-90K \texttt{septuplet} dataset~\cite{xue2019video}, our TMNet can learn to generate the 7-frame high-resolution and high-frame-rate video sequence.
It costs 8.71 days (209.04 hours) to train our TMNet for 600,000 iterations.

In \texttt{Step 2}, we fix the weights of our main network learned in \texttt{Step 1} and only train our TMB block for temporal modulation.
Here, we train our TMNet on the Adobe240fps dataset~\cite{su2017deep}, which has 133 videos in 720P with high-frame-rate (240fps).
At first, We randomly split the Adobe240fps dataset~\cite{su2017deep} into the \texttt{train}, \texttt{val}, and \texttt{test} subsets with 100, 16, and 17 videos, respectively.
Then we split the frames from Adobe240fps \texttt{train}, \texttt{valid}, and \texttt{test} sets into sequences of 7 continuous frames.
We first downsample the original HR frames with the resolution of 1280$\times$720 by a factor of 2 and take them as the ground truths (GTs). 
Then we downsample the GTs to create the corresponding LR input frames by a factor of 4.
All the downsample operations are performed via Bicubic interpolation.
The 1-st and 7-th LR frames of each video sequence are input to our TMNet.
We set the temporal hyper-parameter $t$$\in$$\{\frac{1}{6},\frac{2}{6},\frac{3}{6},\frac{4}{6},\frac{5}{6}\}$ to interpolate 5 intermediate frames.
Supervised by the corresponding 7-frame HR video sequences in Adobe240fps \texttt{test} set as GTs, our TMNet is able to flexibly interpolate intermediate frames according to the temporal hyper-parameter.
It takes 35.26 minutes to train our TMB block for 1,500 iterations on the Adobe240fps dataset~\cite{su2017deep}.

\section{Flexible STVSR with Arbitrary Number of Intermediate Frames}
\label{sec:flexible}

\begin{figure*}[t]
	\centering
	\includegraphics[width=\linewidth]{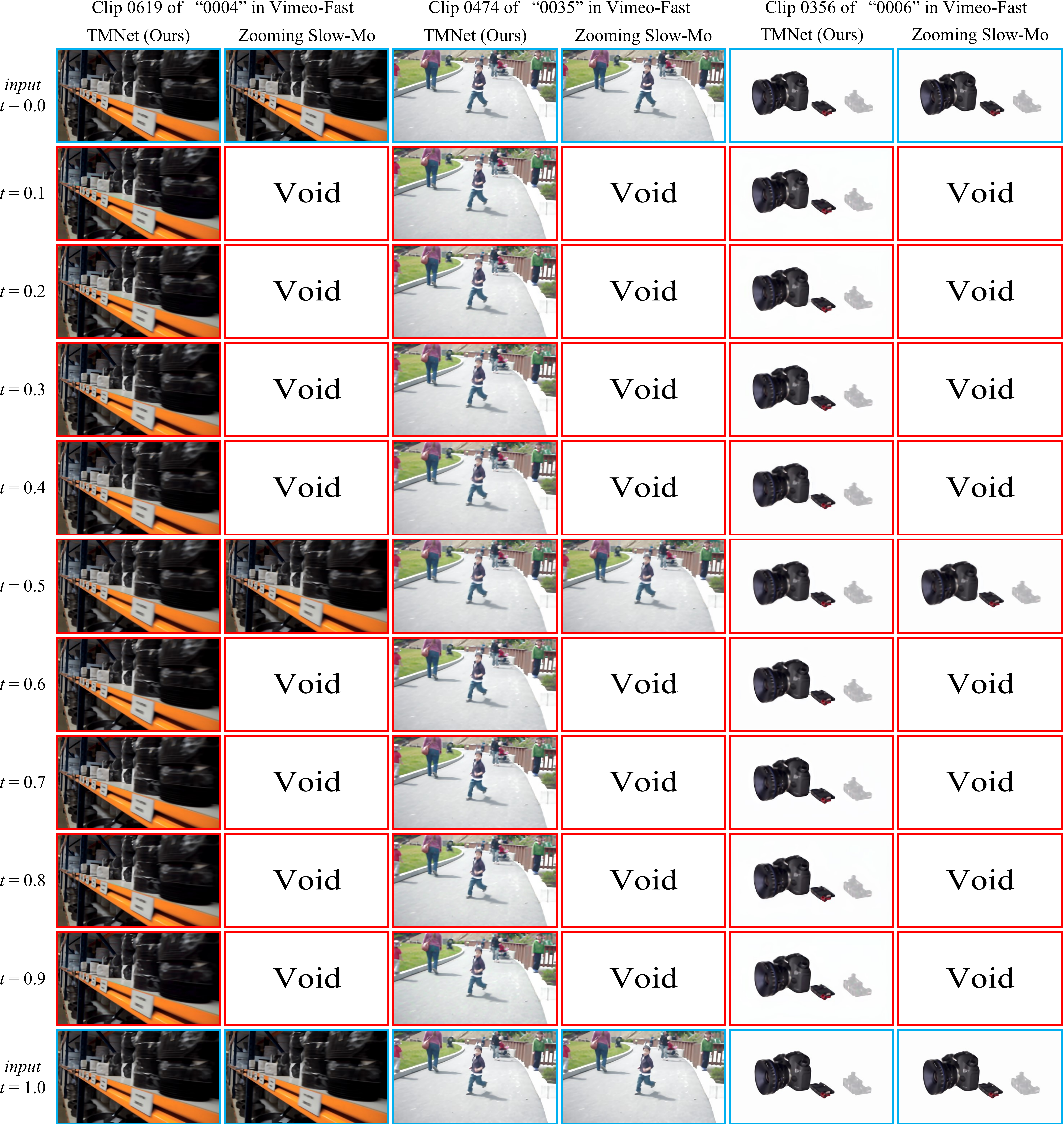}
	\caption{\textbf{Comparison of flexibility on STVSR by our TMNet (1-st, 3-rd, and 5-th columns) and Zooming Slow-Mo~\cite{xiang2020zooming} (2-nd, 4-th, and 6-th columns)} on three video clips from the Vimeo-\texttt{Fast} dataset~\cite{xue2019video}.
	We show the intermediate frames between the adjacent two frames according to the temporal hyper-parameter $t$$\in$$\{0.1,0.2,0.3,0.4,0.5,0.6,0.7,0.8,0.9\}$}
	\label{fig:flexible_STVSR}
\end{figure*}

To show the flexibility of our TMNet for interpolating arbitrary number of intermediate frames on STVSR, we provide the results generated by our TMNet between the input two frames using multiple temporal hyper-parameter $t$.
As the motions in Adobe240fps~\cite{su2017deep} dataset are extremely slow, we validate the flexibility of our TMNet on the Vimeo-90K dataset~\cite{xue2019video}.
To this end, we set the temporal hyper-parameter $t$$\in$$\{0.1,0.2,0.3,0.4,0.5,0.6,0.7,0.8,0.9\}$ to interpolate 9 intermediate frames between any two adjacent frames, though our TMNet is trained to interpolate 5 intermediate frames between Frame 1 and Frame 7. 
The results are shown in Figure~\ref{fig:flexible_STVSR}.
One can see that the interpolated frames vary continuously with the change of $t$ from 0.1 to 0.9.
This demonstrates that our TMNet is feasible to generate a number of intermediate frames, which is different from the training stage. 
That is, our TMNet is very flexible on interpolating arbitrary number of intermediate frames, according to the temporal hyper-parameter $t\in(0,1)$.

\begin{figure*}[t]
	\centering
	\includegraphics[width=\linewidth]{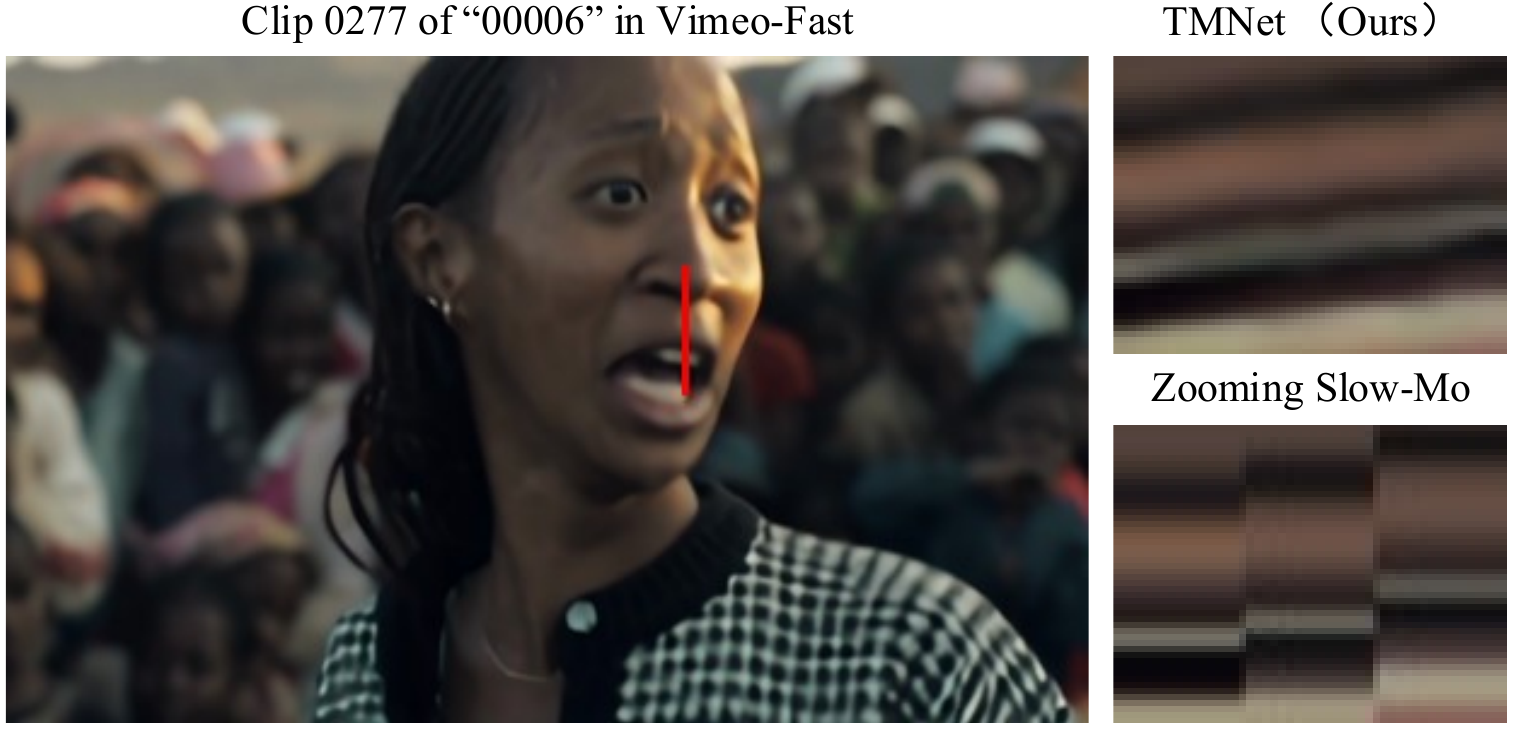}
	\caption{\textbf{Temporal consistency of our TMNet on STVSR}.
	OUr TMNet interpolates 9 frames, while Zooming Slow-Mo~\cite{xiang2020zooming} interpolates 1 frame between Frames 1 and 3.
	We extract a 1D pixel vector over the whole frames from the red line shown in the left figure, and concatenate the 1D pixel vector into a 2D image, which is horizontally scaled to better visualize the temporal consistency of the videos.
	One can see that our TMNet (upper right) achieves clearly consistent temporal interpolation, while Zooming Slow-Mo~\cite{xiang2020zooming} (lower right) suffers from clear breaking variations. 
	}
	\label{fig:consistency}
\end{figure*}

In Figure~\ref{fig:consistency}, we visualize the temporal consistency of our TMNet and Zooming SlowMo~\cite{xiang2020zooming}, on the Clip 0277 of ``00006'' from the Vimeo-\texttt{Fast} set~\cite{xue2019video}.
Our TMNet interpolates 9 frames, while Zooming Slow-Mo~\cite{xiang2020zooming} interpolates 1 frame, between Frames 1 and 3.
To illustrate the temporal motion of the videos, we extract a 1D pixel vector over the whole frames from the red line shown in the left figure, and concatenate the 1D pixel vector into a 2D image.
We observe that our TMNet (Figure~\ref{fig:consistency}, upper right)  produces more consistent temporal motion trajectory than Zooming SlowMo~\cite{xiang2020zooming} (Figure~\ref{fig:consistency}, lower right), which suffers from clear breaking variations. 
%
This demonstrates the superiority of our TMNet on flexible frame interpolation for STVSR.

\section{More Visual Comparisons on STVSR}
\label{sec:comparison}

\begin{figure*}[t]
	\centering
	\includegraphics[width=.8\linewidth]{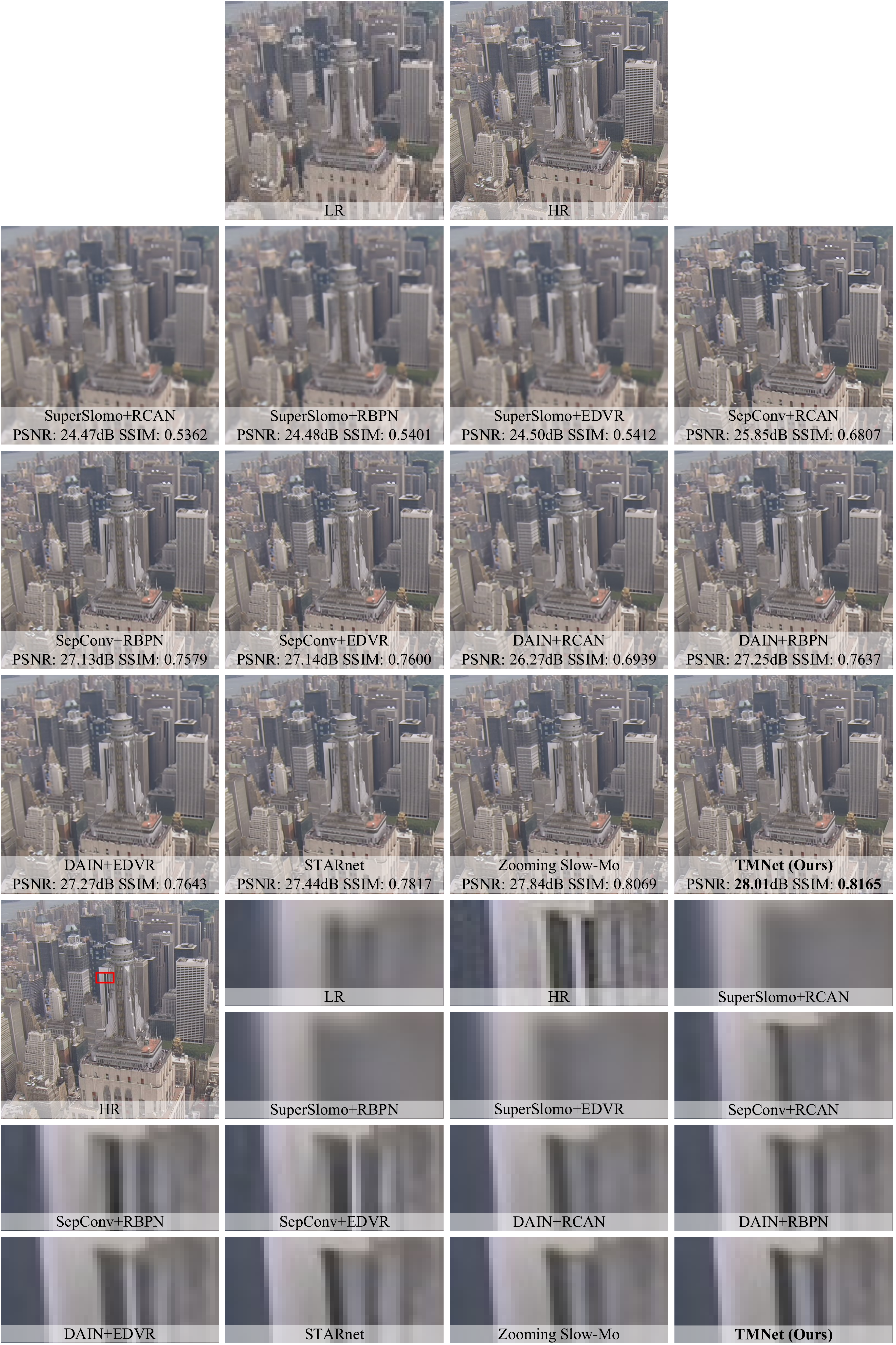}
	\vspace{-2mm}
	\caption{\textbf{Quantitative and qualitative results of our TMNet and other STVSR methods on Clip ``city'' in the Vid4 dataset~\cite{su2017deep}}. 
	For two-stage STVSR methods, we employ SuperSloMo~\cite{jiang2018super}, SepConv~\cite{niklaus2017video} or DAIN~\cite{bao2019depth} for VFI and RCAN~\cite{rcan}, RBPN~\cite{haris2019recurrent} or EDVR~\cite{wang2019edvr} for VSR.
	For one-stage STVSR methods, we compare our TMNet with STARnet~\cite{haris2020space} and Zooming Slow-Mo~\cite{xiang2020zooming}).
	The best results on PSNR (dB) and SSIM~\cite{ssim} are highlighted in \textbf{bold}.
	}
	\label{fig:supplementary_comparison_1}
\end{figure*}

\begin{figure*}[t]
	\centering
	\includegraphics[width=.8\linewidth]{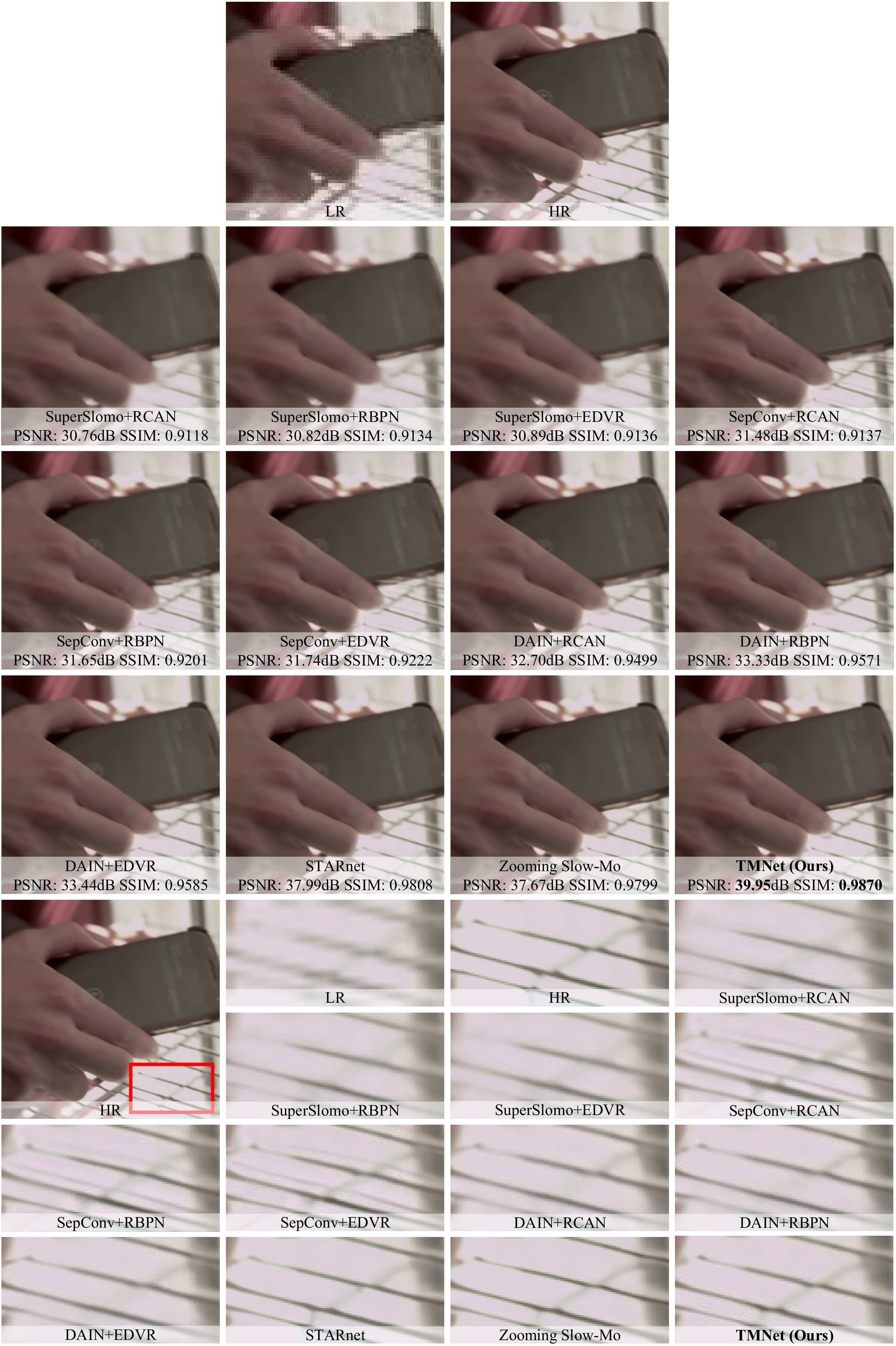}
	\vspace{-2mm}
	\caption{\textbf{Quantitative and qualitative results of our TMNet and other STVSR methods on Clip 0200 of ``00026'' in Vimeo-\texttt{Fast}~\cite{xue2019video}}. 
	For two-stage STVSR methods, we employ SuperSloMo~\cite{jiang2018super}, SepConv~\cite{niklaus2017video} or DAIN~\cite{bao2019depth} for VFI and RCAN~\cite{rcan}, RBPN~\cite{haris2019recurrent} or EDVR~\cite{wang2019edvr} for VSR.
	For one-stage STVSR methods, we compare our TMNet with STARnet~\cite{haris2020space} and Zooming Slow-Mo~\cite{xiang2020zooming}).
	The best results on PSNR (dB) and SSIM~\cite{ssim} are highlighted in \textbf{bold}.
	}
	\label{fig:supplementary_comparison_2}
\end{figure*}

\begin{figure*}[t]
	\centering
	\includegraphics[width=.8\linewidth]{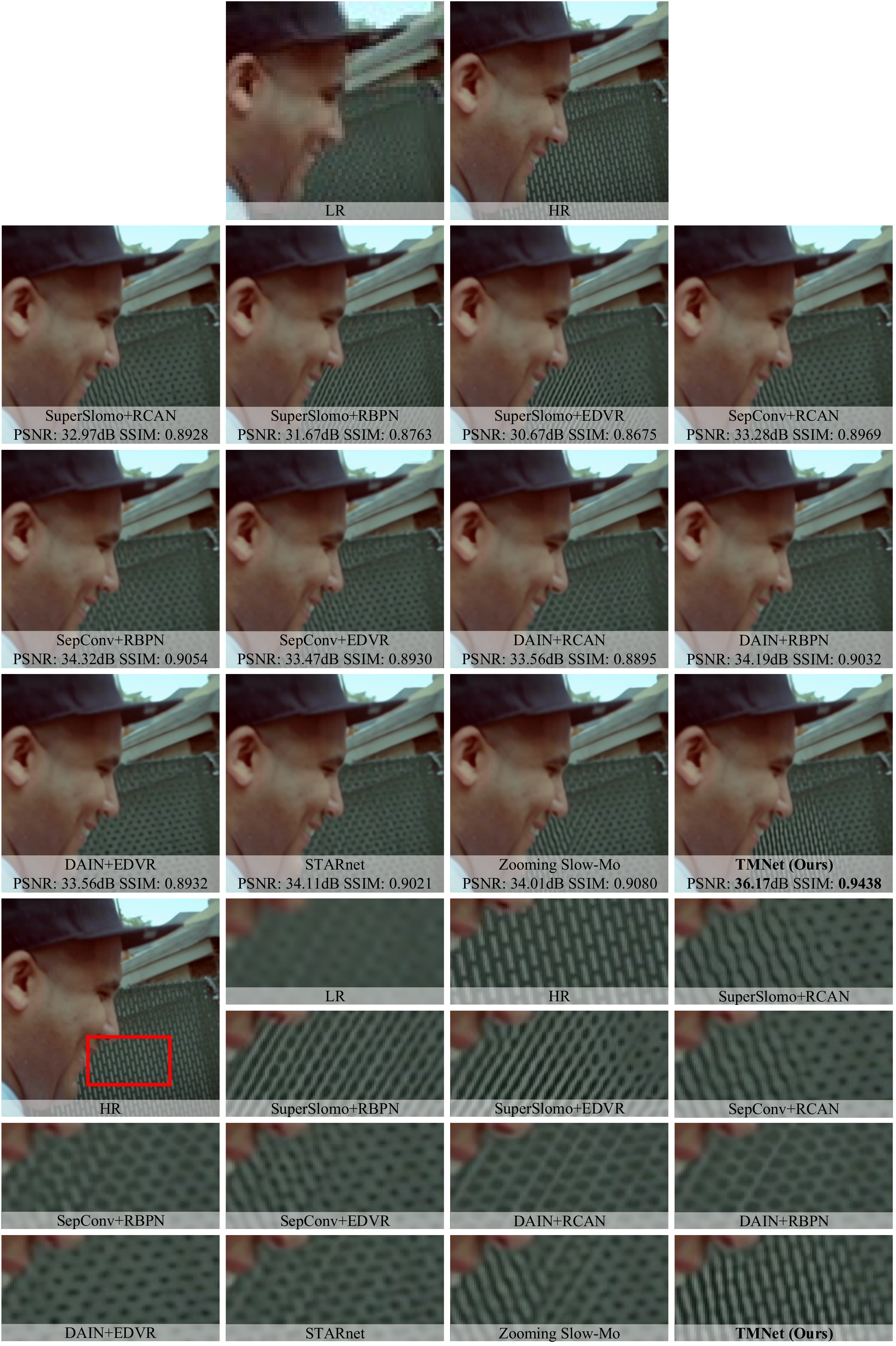}
	\vspace{-2mm}
	\caption{\textbf{Quantitative and qualitative results of our TMNet and other STVSR methods on Clip 0723 of ``00085'' in Vimeo-\texttt{Medium}~\cite{xue2019video}}.
	For two-stage STVSR methods, we employ SuperSloMo~\cite{jiang2018super}, SepConv~\cite{niklaus2017video} or DAIN~\cite{bao2019depth} for VFI and RCAN~\cite{rcan}, RBPN~\cite{haris2019recurrent} or EDVR~\cite{wang2019edvr} for VSR.
	For one-stage STVSR methods, we compare our TMNet with STARnet~\cite{haris2020space} and Zooming Slow-Mo~\cite{xiang2020zooming}).
	The best results on PSNR (dB) and SSIM~\cite{ssim} are highlighted in \textbf{bold}.
	}
	\label{fig:supplementary_comparison_3}
\end{figure*}

\begin{figure*}[t]
	\centering
	\includegraphics[width=.8\linewidth]{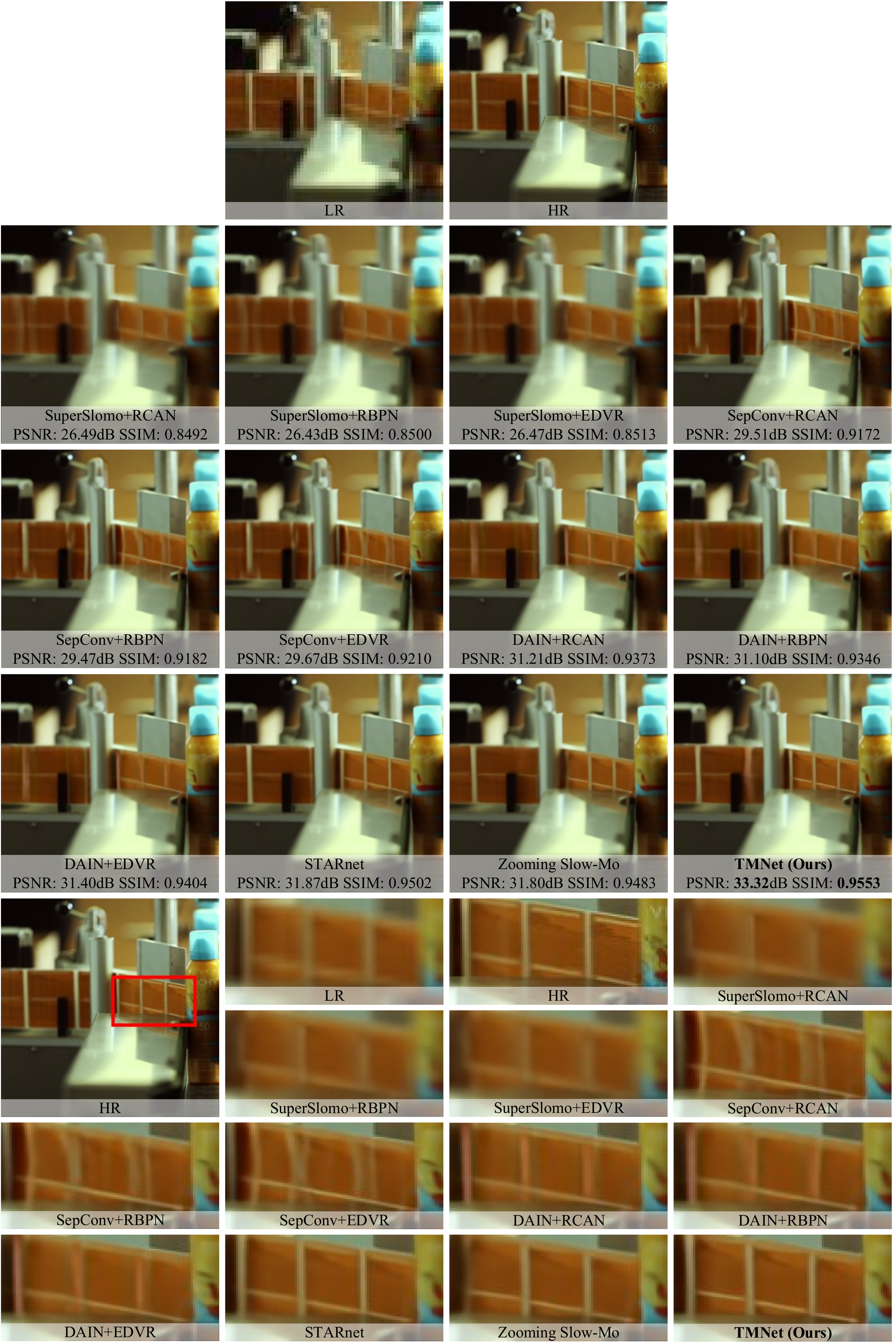}
	\vspace{-2mm}
	\caption{\textbf{Quantitative and qualitative results of our TMNet and other STVSR methods on Clip 0679 of ``00084'' in Vimeo-\texttt{Slow}~\cite{xue2019video}}. 
	For two-stage STVSR methods, we employ SuperSloMo~\cite{jiang2018super}, SepConv~\cite{niklaus2017video} or DAIN~\cite{bao2019depth} for VFI and RCAN~\cite{rcan}, RBPN~\cite{haris2019recurrent} or EDVR~\cite{wang2019edvr} for VSR.
	For one-stage STVSR methods, we compare our TMNet with STARnet~\cite{haris2020space} and Zooming Slow-Mo~\cite{xiang2020zooming}).
	The best results on PSNR (dB) and SSIM~\cite{ssim} are highlighted in \textbf{bold}.
	}
	\label{fig:supplementary_comparison_4}
\end{figure*}

On the Vid4~\cite{vid4} and Vimeo-90K~\cite{su2017deep} datasets, we compare our TMNet with previous one-stage and two-stage STVSR methods.
For one-stage STVSR methods, we compare our TMNet with Zooming SlowMo~\cite{xiang2020zooming} and STARnet~\cite{haris2020space}.
For the two-stage STVSR methods, we perform video frame interpolation (VFI) by SuperSloMo~\cite{jiang2018super}, DAIN~\cite{bao2019depth}, or SepConv~\cite{niklaus2017video}, and perform video super-resolution (VSR) by 
RCAN~\cite{rcan}, RBPN~\cite{haris2019recurrent}, or EDVR~\cite{wang2019edvr}.
We set $t=0.5$ in our TMNet to generate the frame at the middle moment of any two adjacent frames, which means that the 1-st, 3-rd, 5-th, and 7-th LR frames of each clip in Vimeo-90K are fed into our TMNet to reconstruct the 7 HR frames.
All these methods are trained on the Vimeo-90K \texttt{septuplet} dataset~\cite{xue2019video}, and evaluated on the Vimeo-90K \texttt{test} set~\cite{xue2019video} and the Vid4~\cite{su2017deep} dataset.
The visualization results of the comparison result are shown in Figures~\ref{fig:supplementary_comparison_1}-\ref{fig:supplementary_comparison_4}.



\section{Training our TMNet in One-step}
\label{sec:one-stage}
Although trained by a two-step scheme, our TMNet can be directly trained with the proposed TMB block, resulting in a one-step training scheme.
That is, in this one-step scheme, all the parameters of our main TMNet and the TMB block are optimized simultaneously without pre-training.
In our two-step scheme, the two sets of parameters in our main TMNet and the TMB block are optimized separately (first the main TMNet, and then the TMB block).
Here, we compare the performance of our TMNet trained with our two-step and the one-step schemes, resulting in two variants called TMNet-\texttt{two} (the original TMNet) and TMNet-\texttt{one}, respectively.
Both variants are trained on the Adobe240fps \texttt{train} set~\cite{su2017deep} and evaluated on the Adobe240fps \texttt{test} set~\cite{su2017deep}.
As shown in Table~\ref{tab:one_and_two}, comparing with our TMNet-\texttt{two}, the variant TMNet-\texttt{one} suffers from a performance drop of 1.84dB in terms of PSNR, on the Adobe240fps \texttt{test} set~\cite{su2017deep}.
This demonstrates that our TMNet trained in a one-step scheme fail to estimate the motion cues, and interpolate the intermediate frames at an arbitrary moment $t\in(0,1)$.
The main reason is that, in initial training iterations, our TMNet with TMB trained from scratch could not extract useful motion cues from videos, and thus fails to optimize the parameters of our TMB block for meaningful features at an arbitrary moment $t$.

\begin{table}[h]
	\vspace{-2mm}
	\scriptsize
	\centering
	\caption{\textbf{PSNR results of our TMNet trained in two-step or one-step schemes} on Adobe240fps \texttt{test} set~\cite{su2017deep}.}
	\vspace{-3mm}
		\resizebox{\columnwidth}{!}{
			\begin{tabular}{l||cc}
				\Xhline{1pt}
				\rowcolor[rgb]{ .9,  .9,  .9}
				\multicolumn{1}{c||}{\multirow{-1}{*}{Variant}} & TMNet-\texttt{one} & TMNet-\texttt{two}\\
				\hline
				PSNR (dB) & 25.11 & 26.95 \\
				\hline
			\end{tabular}
		}
		\label{tab:one_and_two}
	\vspace{-2mm}
\end{table}

\end{appendices}

\end{document}